\documentclass[10pt,letterpaper]{article}
\usepackage[T1]{fontenc}
\usepackage{lmodern}
\usepackage{arxiv}
\usepackage{amsmath,amssymb}
\usepackage{graphicx,booktabs,array,multirow}
\usepackage{algorithm,algpseudocode}
\usepackage{microtype}
\usepackage{flafter,placeins}
\usepackage[numbers,sort,compress]{natbib}
\usepackage{xurl}
\usepackage[hidelinks,pdfusetitle]{hyperref}

\newcommand{\method}{\textnormal{\textsc{ViCam3D}}}
\newcommand{\basecfg}{\textnormal{\textsc{Base}}}
\newcommand{\covercfg}{\textnormal{\textsc{ViCam-Index}}}
\newcommand{\combinedcfg}[1]{\textnormal{\textsc{#1+ViCam3D}}}
\newcommand{\initcfg}{\textnormal{\textsc{GSC}}}

\title{ViCam3D: Geometry-Supervised Virtual Cameras for Text--3D Scene Retrieval}
\author{%
  Tengjia Kang, Fanghui Sun, Jilong Zhang, He Hao, Xingchen Chen, Bairui Luan, Ying Ma%
  \texorpdfstring{%
    \textsuperscript{*}\\[4pt]
    \normalfont Harbin Institute of Technology\\[2pt]
    \normalfont\small\textsuperscript{*}Correspondence to Ying Ma
    (\href{mailto:y.ma@hit.edu.cn}{\nolinkurl{y.ma@hit.edu.cn}})%
  }{}%
}
\date{}
\renewcommand{\shorttitle}{ViCam3D}

\begin{document}
\raggedbottom
\maketitle

\begin{abstract}
    Text--3D scene retrieval requires bridging the profound modality gap between natural language and unstructured point clouds. While rendering virtual views offers a promising bridge to leverage powerful pretrained image--text models, the retrieval value of these views strictly depends on which scene surfaces they expose. We present \method{} (Virtual Cameras for 3D), a geometry-supervised camera pipeline for bidirectional text--3D scene retrieval. To capture discriminative views without manual annotations, \method{} initializes a budgeted camera set via Greedy Surface Coverage and refines their poses using a lightweight network trained to maximize geometric coverage and observation entropy. By rendering these refined views and encoding them with frozen EVA-CLIP, we construct a reusable visual index that completely bypasses per-scene neural representation fitting. A direction-specific score normalization interface then enables seamless, training-free fusion with existing 3D retrievers. Evaluated on the CrossOver ScanNet and 3RScan validation galleries under a compact 12-view budget, our fused macro Rsum3 reaches 86.38 for text-to-scene and 77.53 for scene-to-text. Furthermore, comprehensive evaluations across seven retriever families reveal a key insight: while ensembling pure 2D models yields redundant visual features, fusing our geometry-aware index with 3D-native retrievers unlocks broad and significant complementary gains. 
\end{abstract}
\section{Introduction}
\label{sec:introduction}

Text--3D scene retrieval connects natural-language descriptions with reconstructed environments, extending language-based access from locating objects within a known scene~\cite{chen2020scanrefer,achlioptas2020referit3d} to searching a complete gallery of 3D spaces. Point clouds are the standard and most direct output of 3D scene reconstruction, enabling accurate, language-driven retrieval over point cloud galleries is a critical task. However, bridging the modality gap between unstructured 3D points and natural language remains challenging. Leveraging powerful pretrained image--text models for point cloud retrieval requires constructing virtual 2D observations that comprehensively capture the 3D scene's geometric and visual content.

Existing retrieval methods connect language and 3D content through various representations. Native point-cloud--text matchers learn 3D scene-level correspondence directly~\cite{roma,feng2025interactive}, while other approaches rely on original scene photographs~\cite{whereami}, RGB--pointmaps~\cite{uniscene3d}, or rendered views from fitted neural graphics~\cite{retri3d}. Rendered observations have proven particularly effective for 3D visual understanding~\cite{openins3d}, allowing models to bypass the sparsity and irregularity of point clouds. However, this raises a central question for point cloud retrieval: how can a finite set of virtual cameras automatically and optimally expose the most discriminative visual evidence from an unstructured 3D scene?

For scene retrieval, camera placement dictates which 3D surfaces are encoded into the gallery index and made available to the image--text model. Indoor occlusion, incomplete geometry, and overlapping fields of view often yield empty or highly redundant images. Consequently, simply rendering a fixed number of heuristic views does not guarantee sufficient scene coverage for accurate retrieval. Effective virtual camera construction must identify valid viewpoints, distribute observations across visible surfaces, and refine their poses within a strict view budget. Because this process relies on geometric visibility, it provides query-independent supervision, allowing the full retrieval index to be prepared offline before queries arrive.

We propose \method{} (Virtual Cameras for 3D), a geometry-supervised virtual camera pipeline designed for bidirectional text--point-cloud retrieval. From point coordinates, the pipeline estimates floor support, filters camera centers, and enumerates viewing directions. First-hit ray traversal identifies the surface cells visible from each candidate. Greedy Surface Coverage (\initcfg{}) applies the standard maximum-coverage principle~\cite{nemhauser1978submodular} to initialize cameras that expose complementary surfaces. A lightweight network then predicts bounded XYZ, yaw, and pitch updates from geometry features and pooled scene context. Its supervision combines surface coverage with normalized observation entropy. Sequential acceptance retains valid proposals that improve the joint objective, and two refinement rounds update the camera set before rendering.

Surfel rendering converts the accepted views into images, which frozen EVA-CLIP~\cite{evaclip} directly encodes into a reusable scene index without requiring per-scene neural fitting. During retrieval, direction-specific score standardization~\cite{markov2012normalization,louis2025know} allows the resulting image--text similarities to be seamlessly fused with any existing 3D retriever, preserving its original checkpoint and native representation.

We evaluate \method{} on the CrossOver ScanNet and 3RScan validation galleries. Under a compact 12-view budget, the fused seven-family macro Rsum3 reaches an impressive 86.38 for text-to-scene and 77.53 for scene-to-text retrieval. Notably, all 48 paired configurations consistently improve over their base retrievers. Family-wise sweeps separate three regimes: visual-dominated gains for weak image--text architecture transfers, largely redundant ensembling of frozen CLIP2CLIP with EVA-CLIP, and broad complementary gains for RoMa, IDeal Source, and CrossOver. Matched-pool ablations isolate camera initialization, while supervision, pose, acceptance, and refinement-round studies examine the learned camera policy. Budget and timing analyses quantify the preparation, storage, and retrieval costs. Our contributions are:

\begin{itemize}
\item A geometry-supervised virtual camera pipeline for text--point-cloud retrieval that combines visible-surface coverage initialization with learned position and orientation refinement, dynamically capturing discriminative views without manual camera annotations.
\item A reusable visual index built directly from raw colored point clouds and frozen image--text models. By completely bypassing per-scene neural representation fitting, it enables highly efficient offline indexing, and features a direction-specific score interface for seamless, training-free fusion with existing 3D retrievers.
\item A comprehensive empirical evaluation across seven retriever families. We systematically isolate the benefits of learned camera refinement and demonstrate that while ensembling pure 2D models (e.g., CLIP) yields redundant features, fusing our geometry-aware visual index with 3D-native retrievers unlocks broad and significant complementary gains.
\end{itemize}

\section{Related Work}
\label{sec:related_work}
\paragraph{Text--3D scene retrieval.}
RoMa learns point-cloud--text matching with attention and robust negative contrastive learning~\cite{roma}, and IDeal refines retrieval through question--answer interaction~\cite{feng2025interactive}.
CrossOver aligns multiple modalities at the instance and scene levels and supports point-cloud--text retrieval with modality-specific encoders~\cite{crossover}.
Text2SceneGraphMatcher matches language-derived graphs to 3D scene graphs~\cite{whereami}.
LangLoc combines graph matching with CLIP scores to retrieve a scene before estimating the observer's position and heading~\cite{langloc}.

\paragraph{Visual indexing of scenes.}
CLIP2CLIP ranks scenes by the highest CLIP similarity between a query and a scene image~\cite{whereami,clip}.
POMA-3D retrieves scenes from multi-view pointmaps~\cite{mao2026poma3d}, and UniScene3D jointly encodes RGB images and pixel-aligned world-coordinate pointmaps for zero-shot scene retrieval~\cite{uniscene3d}.
Retri3D indexes fitted neural graphics representations, including NeRFs and 3D Gaussian splats, with features from views selected to reduce rendering artifacts~\cite{retri3d}.
OpenIns3D uses point-cloud renderings to label 3D instances~\cite{openins3d}.
EVA-CLIP provides a pretrained image--text encoder for transferable visual representation~\cite{evaclip}.
The visual branch of \method{} uses the frozen encoder to build a reusable full-gallery index from surfel renderings of colored point clouds.

\paragraph{Camera selection and pose refinement.}
CrossOver samples recorded views by separation in camera position and orientation~\cite{crossover}.
Video-3D LLM greedily selects RGB-D frames by voxel coverage~\cite{zheng2025video3d}.
Sun et al.\ optimize camera poses using a learned coverage score~\cite{sun2021viewselection}, and MVTN learns task-specific viewpoints for shape recognition~\cite{hamdi2021mvtn}.
CONVERGE prioritizes views of under-observed geometry for radiance-field reconstruction~\cite{chen2026converge}.
Greedy marginal-gain selection is the standard approximation for cardinality-constrained monotone submodular maximization~\cite{nemhauser1978submodular}.
\method{} applies this principle to first-hit visible surface cells to initialize a fixed budget of virtual cameras.
A geometry-only teacher based on surface coverage and observation entropy supervises refinement of camera position, yaw, and pitch.

\paragraph{Score normalization and fusion.}
Z-score normalization is established for combining retrieval scores~\cite{markov2012normalization,louis2025know}.
Querybank Normalisation uses an auxiliary query bank to address hubness~\cite{qbnorm}, and Dual Bank Normalization uses both query and gallery banks~\cite{dbnorm}.
For each query in either retrieval direction, \method{} standardizes each branch's scores across the current candidates and combines the standardized scores by weighted sum.

\section{Method}
\label{sec:method}

\subsection{Overview}
\label{sec:method_overview}
\begin{figure}[!htbp]
\centering
\includegraphics[width=\textwidth]{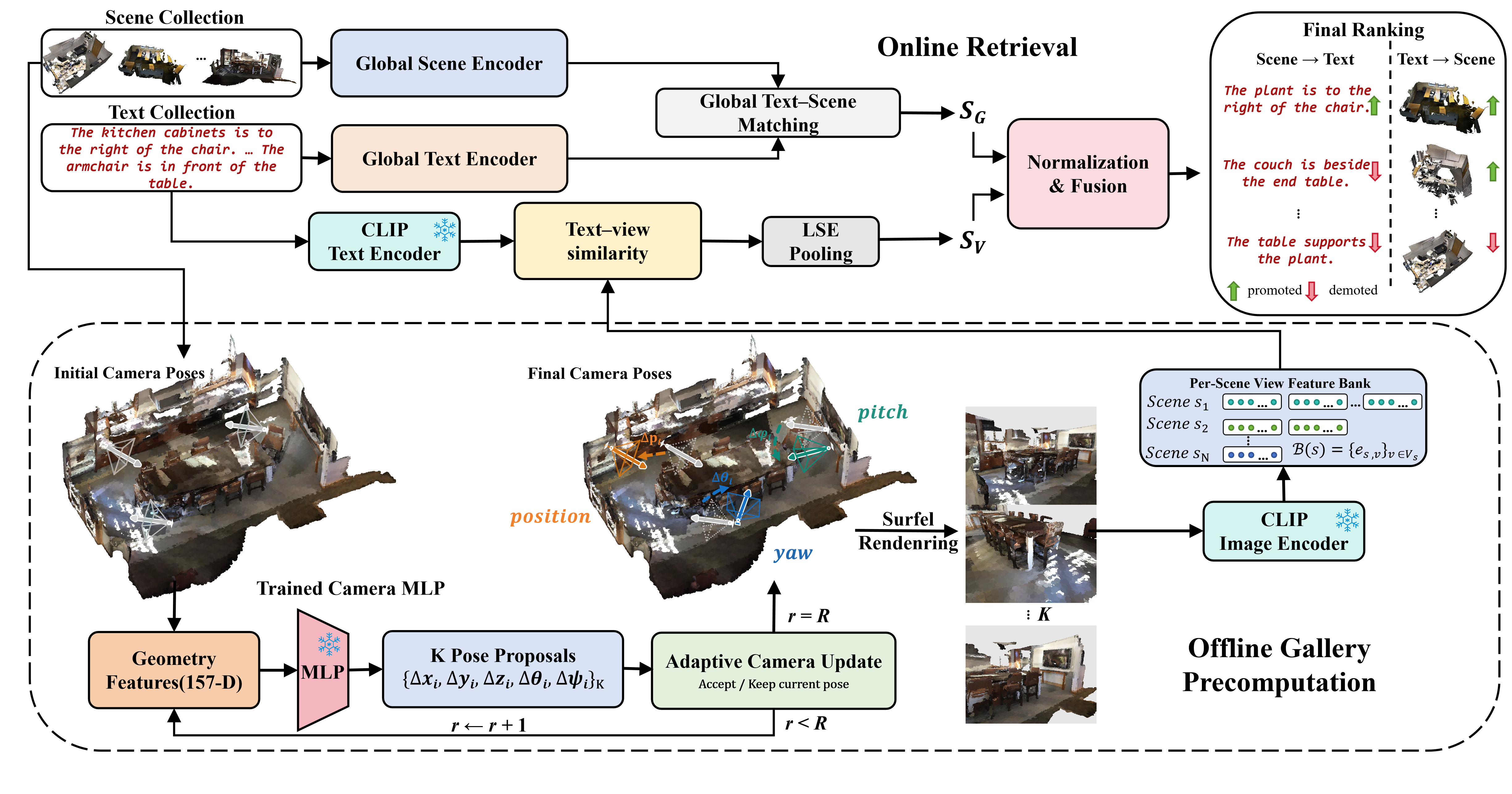}
\caption{\textbf{Overview of \method{}.} The dashed lower block performs offline gallery precomputation. Greedy Surface Coverage (\initcfg{}) supplies the initial camera poses. Geometry Features (157-D), the trained Camera MLP, $K$ Pose Proposals, and Adaptive Camera Update implement geometry-supervised camera refinement over $R$ rounds. Surfel Rendering and the frozen EVA-CLIP Image Encoder form the Per-Scene View Feature Bank $\mathcal B(s)$. Online, the frozen EVA-CLIP Text Encoder, Text--View Similarity, and normalized LSE Pooling produce $S_V$. A frozen global retriever performs Global Text--Scene Matching to obtain $S_G$. Direction-specific Normalization \& Fusion yields the bidirectional final rankings. Snowflakes mark modules frozen during gallery construction and retrieval.}
\label{fig:overview}
\end{figure}

Figure~\ref{fig:overview} divides \method{} into offline gallery construction and online retrieval. Offline, \initcfg{} initializes the camera poses, and a geometry-supervised MLP refines them through sequential updates. Surfel rendering and a frozen image encoder produce a reusable per-scene feature bank. Online, a frozen text encoder and normalized log-sum-exp pooling produce the visual score $S_V$. Direction-specific normalization combines this score with an existing retriever's score $S_G$. We denote the standalone visual index by \covercfg{}, the existing retriever by \basecfg{}, and their combination by the named retriever+\method{}.

\subsection{Problem Formulation}
\label{sec:problem_formulation}
Let $\mathcal D=\{(q_i,s_i)\}_{i=1}^{L}$ pair descriptions $\mathcal Q=\{q_i\}_{i=1}^{L}$ with reconstructed scenes $\mathcal S=\{s_i\}_{i=1}^{L}$, where $q_i$ describes $s_i$. Pairwise scores $M_{ij}=S(q_i,s_j)$ form a raw compatibility matrix $M\in\mathbb R^{L\times L}$. Text-to-scene retrieval ranks scenes by row $M_{i,:}$ for $q_i$, with target $s_i$; scene-to-text retrieval ranks descriptions by column $M_{:,j}$ for $s_j$, with target $q_j$. Both directions reuse raw evidence and recompute candidate-axis normalization and ranking (Sec.~\ref{sec:fusion}).

Scene $s_j$ has a metric, Z-up colored point cloud $P_{s_j}=\{(x_n,a_n)\}$, with $x_n\in\mathbb R^3$ and RGB $a_n$. From $P_{s_j}$, the visual branch derives normals, occupancy, visibility, and a query-independent set $V_{s_j}$ of at most $K$ cameras $v=(A_v,R_v,t_v)$; $A_v$ denotes intrinsics and $(R_v,t_v)$ the pose. These cameras expose complementary reconstructed surfaces and yield a reusable visual index for both directions.

\subsection{Candidate Camera Generation}
\label{sec:camera_generation}
Local XYZ geometry estimates horizontal floor support and constructs an admissible grid of camera centers using support and observed-point-clearance constraints. Deterministic farthest-point selection retains up to six positions. At each position, eight azimuths spaced by $45^\circ$ combine with pitch angles of $-30^\circ$, approximately $-8.53^\circ$, and $15^\circ$, yielding at most 144 candidates in position--azimuth--pitch order. Intrinsics use $640\!\times\!480$ images and focal lengths of 420 pixels. Appendix~\ref{sec:camera_implementation_conventions} derives the floor estimate, admissible centers, and exact orientation set from the fixed protocol constants.

These rules construct virtual camera poses directly from the reconstruction. Observed-point clearance measures separation from reconstructed geometry; free space behind missing geometry remains unknown.

\subsection{Greedy Surface Coverage Initialization}
\label{sec:camera_initialization}
\label{sec:surface_method}
Initialization is geometry-only and query-independent. We voxelize $P_s$ at $h_{\rm occ}$; each candidate casts $N_{\rm rays}=160\!\times\!120=19{,}200$ rays on a fixed raster. $H_v$ stores one first occupied-cell hit per successful ray, so $|H_v|/N_{\rm rays}$ is the ray-hit fraction. We discard views below $\rho_{\min}$ and quantize hits at $h_{\rm cov}$ into the unique visible set $C_{s,v}$. Appendix Eq.~\ref{eq:visibility_definition} defines the exact traversal and depth.

For eligible candidates $\mathcal P_s$ and budget $K$, define the optimal coverage value as
\begin{equation}
F_{cov}^{*}(s,K) = \max_{V \subseteq \mathcal{P}_s, \vert{}V\vert{} \le K} \left\vert{} \bigcup_{v\in V}C_{s,v} \right\vert{}
\label{eq:coverage}
\end{equation}


\begin{samepage}
Starting from $V_0=U_0=\varnothing$, the standard marginal-gain greedy update~\cite{nemhauser1978submodular} selects
\begin{equation}
v^\star=\arg\max_{v\in\mathcal P_s\setminus V_{t-1}}
|C_{s,v}\setminus U_{t-1}|.
\label{eq:greedy}
\end{equation}
\end{samepage}
where $U_{t-1}=\bigcup_{v\in V_{t-1}}C_{s,v}$ and $V_t=V_{t-1}\cup\{v^\star\}$. The rule favors new coverage over total visible area, with ties broken by candidate order. Algorithm~\ref{alg:c4_initialization} details the complete procedure, termed Greedy Surface Coverage (\initcfg{}), which outputs the ordered poses $V_s^{(0)}=(v_1,\ldots,v_K)$. The objective targets geometric observation diversity, while our retrieval experiments measure the resulting language relevance.

\begin{algorithm}[!htbp]
\caption{\textbf{Greedy Surface Coverage (\initcfg{}) camera initialization.} Fixed grids and first-hit rays define coverage and depth. \textsc{SavedPad} pads to $K$ by fixed scene-seeded pseudorandom draws with replacement.}
\label{alg:c4_initialization}
\begin{algorithmic}[1]
\Require Ordered candidates $\mathcal P_s^{\rm cand}$, cloud $P_s$, budget $K$; fixed $h_{\rm occ},h_{\rm cov},\rho_{\min}$ and $N_{\rm rays}=160\!\times\!120$
\Ensure Ordered initial poses $V_s^{(0)}$
\State $O_s\gets\Call{Voxelize}{P_s,h_{\rm occ}}$
\State $\mathcal P_s\gets\varnothing$
\For{$v\in\mathcal P_s^{\rm cand}$ in candidate order}
    \State $H_v\gets\Call{FirstHit}{O_s,v}$ \Comment{one entry per hit ray}
    \If{$|H_v|/N_{\rm rays}\ge \rho_{\min}$}
        \State $C_{s,v}\gets\Call{UniqueCells}{H_v,h_{\rm cov}}$
        \State $\mathcal P_s\gets\mathcal P_s\cup\{v\}$
    \EndIf
\EndFor
\State \textbf{fail if} $\mathcal P_s=\varnothing$
\State $V\gets(),\quad U\gets\varnothing$
\While{$|V|<\min(K,|\mathcal P_s|)$}
    \State $v^\star\gets\displaystyle\arg\max_{v\in\mathcal P_s\setminus V}|C_{s,v}\setminus U|$
    \State $V\gets\Call{Append}{V,v^\star}$
    \State $U\gets U\cup C_{s,v^\star}$
\EndWhile
\If{$|V|<K$}
    \State $V\gets\Call{SavedPad}{V,\mathcal P_s,K,s}$
\EndIf
\State \Return $V_s^{(0)}\gets V$
\end{algorithmic}
\end{algorithm}

Dataset-specific view budgets, ablation protocols, and the padding convention are specified in Sec.~\ref{sec:setup} and Appendix~\ref{sec:point_geometry}.

\subsection{Geometry-Supervised Camera Refinement}
\label{sec:camera_refinement}
The camera network refines the \initcfg{} poses $V_s^{(0)}$. Each valid camera slot has a 157-dimensional geometry descriptor comprising pooled depth statistics, ray-hit support, cross-view novelty, the current pose, observed-point clearance, and scene extent. A shared MLP embeds each slot, concatenates its embedding with the masked mean context of all valid slots, and predicts five tanh-normalized updates $(\Delta x,\Delta y,\Delta z,\Delta\psi,\Delta\theta)$. Each yaw or pitch update is bounded by $10^\circ$. Appendix Eq.~\ref{eq:pose_unnormalization} maps all five normalized outputs to physical offsets and applies the cumulative pose bounds; roll and intrinsics remain fixed. Appendix~\ref{sec:supp_camera_lambda_sweep} specifies the network layers.

\begin{figure}[!htbp]
\centering
\includegraphics[width=.82\linewidth]{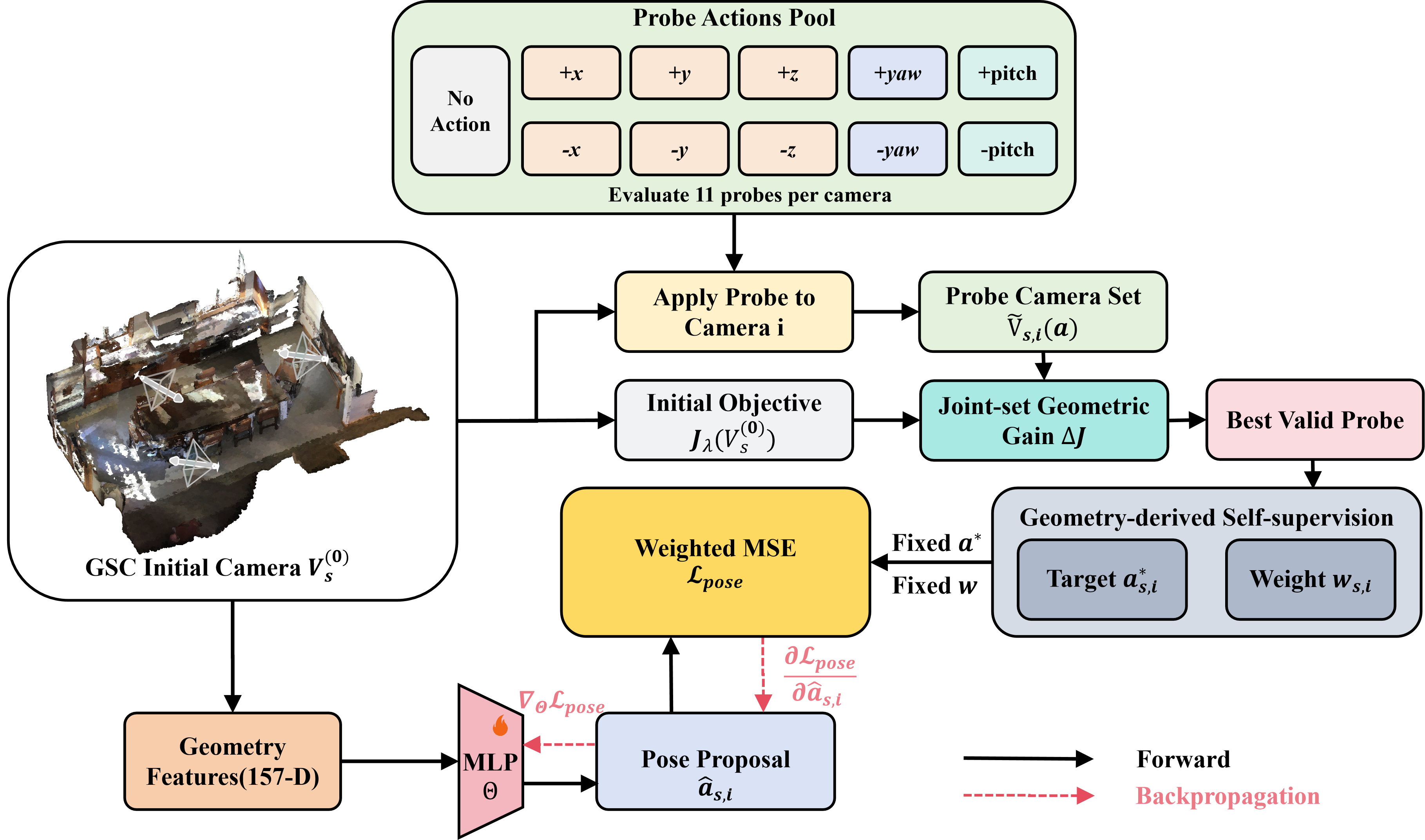}
\caption{\textbf{Geometry-supervised camera-MLP training.} For each \initcfg{} camera, eleven probes replace its pose while the other initial poses stay fixed. Joint-objective gains against $J_\lambda(V_s^{(0)})$ select the best valid target $\mathbf a_{s,i}^{*}$ and determine $w_{s,i}$. The MLP predicts $\widehat{\mathbf a}_{s,i}$ from 157-dimensional geometry features. Fixed targets and weights enter $\mathcal L_{\rm pose}$ (Eq.~\ref{eq:camera_pose_loss}). Black arrows show forward computation, and red dashed arrows show gradients through the pose prediction to MLP parameters $\Theta$.}
\label{fig:camera_mlp_training}
\end{figure}

Training uses a geometry-only teacher (Fig.~\ref{fig:camera_mlp_training}). Quantizing the centers of all occupied voxels at resolution $h_{\rm cov}$ gives a fixed universe of $N_{\mathrm{cell}}$ surface cells for each scene. For a camera set $V$, its coverage fraction is $C(V)=|\bigcup_{v\in V}C_{s,v}|/N_{\mathrm{cell}}$. Let $n_u(V)$ be the pooled first-hit ray count for surface cell $u\in\{1,\ldots,N_{\mathrm{cell}}\}$ and $n_0(V)$ the background count. With $p_u(V)=n_u(V)/\sum_{\ell=0}^{N_{\mathrm{cell}}}n_\ell(V)$, the observation entropy is $H(V)=-\sum_{u=0}^{N_{\mathrm{cell}}}p_u(V)\log p_u(V)$, taking $0\log0=0$. Both teacher targets and inference-time gains use
\begin{equation}
J_\lambda(V)=(1-\lambda)C(V)+\lambda\frac{H(V)}{\log(N_{\mathrm{cell}}+1)}.
\label{eq:camera_objective}
\end{equation}
The main configuration uses $\lambda=0.50$.

For each initial slot, the teacher evaluates 11 actions comprising zero motion and positive or negative steps along each of the five predicted axes, at 80\% of their bounds. Each probe replaces only that slot in $V_s^{(0)}$. The best valid positive-gain action supplies the normalized target $\mathbf a_{s,i}^{*}$. The target is zero when no action improves the objective. Its joint-objective gain is $g_{s,i}\geq0$. Let $K_s\leq K$ denote the number of valid, non-padded camera slots for scene $s$. With scene mean $\bar g_s=K_s^{-1}\sum_{i=1}^{K_s}g_{s,i}$, the gain weight is
\begin{equation}
w_{s,i}=1+\min\!\left(\frac{g_{s,i}}{\max(\bar g_s,10^{-12})},4\right).
\label{eq:camera_gain_weights}
\end{equation}
The MLP outputs a five-dimensional normalized prediction for XYZ, yaw, and pitch, denoted by $\widehat{\mathbf a}_{s,i}\in[-1,1]^5$. We minimize the gain-weighted displacement loss
\begin{equation}
\mathcal L_{\rm pose}=\frac{1}{5B}\sum_{s=1}^{B}
\frac{\sum_i w_{s,i}\left\|\widehat{\mathbf a}_{s,i}-\mathbf a_{s,i}^{*}\right\|_2^2}
{\sum_i w_{s,i}}.
\label{eq:camera_pose_loss}
\end{equation}
Both inner sums run over these $K_s$ slots, and $B$ is the number of scenes in the batch. The factor $1/5$ averages errors over the five normalized components, and the outer mean weights scenes equally. Tensor padding is masked in context pooling and the loss. AdamW updates the MLP through $\mathcal L_{\rm pose}$, with teacher targets and gain weights fixed.

At refinement round $r\in\{1,\ldots,R\}$, the MLP predicts offsets from $V_s^{(r-1)}$ and produces the accepted pose sequence $V_s^{(r)}$, with $V_s^{(0)}$ the \initcfg{} initialization and $V_s=V_s^{(R)}$. We clip each proposal to the cumulative bounds relative to its initial \initcfg{} pose. In \initcfg{} slot order, we replace the current pose if the placement is geometrically valid, the first-hit fraction is at least 0.10, and the replacement increases the joint $J_\lambda$ by more than $10^{-12}$. Each decision uses the poses already accepted in that round. A rejected slot keeps its current pose. Figure~\ref{fig:overview} names this sequential accept-or-keep procedure Adaptive Camera Update.

Before the next round, we recompute the geometry features and the scene context from the accepted set. The main comparison uses two rounds (Sec.~\ref{sec:camera_rounds}). We render the final sequence $V_s$ once during query-independent index construction (Fig.~\ref{fig:overview}). The fixed control sets $V_s=V_s^{(0)}$.

\subsection{Virtual Observation Index}
\label{sec:aggregation}
We render the accepted camera sequence from XYZ/RGB using Surfel splats with estimated normals. Frozen EVA02-E-14-plus~\cite{evaclip} supplies unit image and text features,
\begin{equation}
e_{s,v}=\operatorname{norm}(E_{\mathrm{img}}(R_{s,v})),\quad
t_q=\operatorname{norm}(E_{\mathrm{text}}(q)),
\end{equation}
where $R_{s,v}$ is the rendered RGB image. We retain native $224\!\times\!224$ preprocessing, 1,024-dimensional features, and 77-token text context. Each branch receives the full query string and applies its native context limit. Camera construction, rendering, and image encoding occur once per scene. The resulting feature bank $\mathcal B(s)=\{e_{s,v}\}_{v\in V_s}$ is reused across queries; online computation produces $t_q$ and its similarities to the cached descriptors.

Cosine similarity $c_v(q,s)=\langle t_q,e_{s,v}\rangle$ scores each observation. Normalized log-sum-exp (LSE), implemented stably with $\tau=0.03$, gives the visual score $S_V$:
\begin{equation}
S_V(q,s)=\tau\log\!\left[
\frac{1}{|V_s|}\sum_{v\in V_s}\exp(c_v(q,s)/\tau)\right].
\label{eq:lse}
\end{equation}
The normalization makes the score invariant to duplicating the complete observation set, while selectively repeated views retain their relative weight. At small $\tau$, LSE emphasizes the most compatible observations while retaining support from multiple views. Sorting $S_V$ over the complete gallery provides standalone visual retrieval. With feature dimension $d$, scoring and descriptor storage scale as $O(dKL)$; camera construction, rendering, and encoding incur separate offline costs.

\subsection{Integration with Existing Scene Retrievers}
\label{sec:fusion}
An existing retriever supplies $S_G(q,s)=r_\theta(q,s)$ through its native scene representation and text encoder. Its checkpoint and scores remain frozen during fusion, which requires no additional fitting. Each base method retains its original training and input history (Sec.~\ref{sec:setup}).

Using established score standardization~\cite{markov2012normalization,louis2025know}, we compute population statistics for each branch $X\in\{G,V\}$ over the complete candidate set in the current retrieval direction and form the main combination as
\begin{align}
Z_X(q,s)&=\frac{S_X(q,s)-\mu_X(q)}{\sigma_X(q)+10^{-12}},
\label{eq:zscore}\\
S_{G+V}(q,s)&=(1-\gamma)Z_G(q,s)+\gamma Z_V(q,s).
\label{eq:fusion}
\end{align}
For text-to-scene retrieval, each row is normalized across all gallery scenes for one text query. For scene-to-text retrieval, we transpose the raw score matrices and normalize across all candidate texts for one scene query. Each direction therefore uses its own statistics.

We fix $\gamma=0.75$ in the main comparison after observed-validation exploration and share it across methods, text towers, datasets, and directions. Centering adds a query-constant term, so ranking is equivalent to raw-score mixing with coefficients $(1-\gamma)/(\sigma_G+10^{-12})$ and $\gamma/(\sigma_V+10^{-12})$. A constant row contributes zero. These gallery-dependent statistics standardize score dispersion across candidates.

All query--candidate pairs require finite scores, and every scene requires eligible visual observations. Lexical candidate IDs resolve ties. The appendix reports the shared-weight selection and the current ablation protocols.

\section{Experiments}
\label{sec:experiments}

\subsection{Experimental Setup}
\label{sec:setup}

\paragraph{Datasets and retrieval protocol.}
We use the official CrossOver splits of ScanNet~\cite{dai2017scannet} and 3RScan~\cite{wald2019rio,crossover}, retaining scenes with both point clouds and referrals. The training/validation splits contain 1,151/291 and 1,073/149 scenes, respectively. Each validation scene contributes one query and one gallery candidate. Following CrossOver, we sample up to ten SceneVerse pairwise relational referrals~\cite{jia2024sceneverse} per scene without replacement, using a fixed seed and preserving split order. All branches receive the same resulting text and apply their native tokenizers. Query hashes are retained; exact string identity with the authors' release is unverified.

ScanNet reconstructed vertices and 3RScan OBJ/MTL textures supply XYZ/RGB. Camera construction and visibility use XYZ; Surfel rendering also uses RGB and estimated normals. We use reconstruction coordinates directly, without ground-truth axis alignment or instance filtering. Appendix~\ref{sec:point_geometry} documents geometry compatibility cases and repeated view slots.

\paragraph{Camera policy and visual retrieval.}
The main comparison uses $K=12$ on both datasets and two rounds of five-dimensional camera refinement with $\lambda=0.50$. We train three camera seeds for 30 epochs on 128 geometry-only fitting scenes, retain 32 separate diagnostic scenes, and use the final-epoch checkpoints. All fitting and evaluation scenes in this comparison use the same view budget. \covercfg{} uses frozen EVA02-E-14-plus, normalized LSE with $\tau=0.03$, and the native 77-token context~\cite{evaclip}. The initialization, supervision, pose, and acceptance ablations use ScanNet $K=12$ and 3RScan $K=6$; their full protocols appear in the appendix.

\paragraph{Base retrievers and training.}
We restrict the main comparison to the shared colored point-cloud galleries. The native UniScene3D pipeline requires additional RGB--pointmap pairs~\cite{uniscene3d}, and Retri3D relies on neural rendering of fitted scene representations~\cite{retri3d}. We evaluate seven \basecfg{} families. RoMa~\cite{roma}, IDeal Source~\cite{feng2025interactive}, ESA~\cite{zhu2023esa}, CRCL~\cite{qin2023crcl}, and VSE$\infty$~\cite{chen2021vseinfty} each use BERT and BiGRU, yielding ten fresh joint fits with one fixed training seed and native presets. These families share frozen DGCNN descriptors computed from XYZ using 200 patches of 1,024 points per scene. IDeal Source uses the fixed query without interactive enrichment. Checkpoints maximize equal-dataset validation text-to-scene Rsum3, with earliest-epoch tie breaking. Completed schedules span 13--300 epochs; Appendix~\ref{sec:joint_training} lists the presets and selected epochs.

Text2SceneGraphMatcher and LangLoc are not included as base rows. Both assume semantic instance segmentation and typed 3D scene graphs, which fall outside the shared colored point-cloud input condition; their published retrieval evaluations also use ScanScribe/Human queries and gallery definitions rather than the CrossOver full galleries~\cite{whereami,langloc}. LangLoc further treats scene retrieval as the first stage of pose-indexed localization and dialog-based disambiguation. Adapting either method would therefore change the input information and evaluation task rather than yield a matched base-retrieval comparison.

CLIP2CLIP~\cite{whereami} retains frozen OpenAI CLIP ViT-B/32 and native scoring on fixed \initcfg{} Surfel banks. Because its source CLIP is neither fine-tuned nor retrained, it is training-free---operationally zero-shot with respect to target labels. CrossOver uses the official same-split checkpoint and native BLIP text and point towers, retaining its auxiliary RGB/CAD training exposure~\cite{crossover}. All base scores remain fixed in the fusion comparison. Thus, base-to-fusion gains measure the combined addition of reconstructed color and visual pretraining; matched-pool ablations isolate camera initialization.

\paragraph{Metrics and aggregation.}
We report R@1 and Rsum3=$\mathrm{R@1}+\mathrm{R@5}+\mathrm{R@10}$ in percentage points and MRR on a 0--1 scale. Rankings use the complete gallery with lexical ID tie breaking. Camera seeds are ranked separately before their metrics are averaged. Macro results weight datasets equally; seven-family means first average text towers within each family, then weight families equally. The visual-only reference is shared across families. Reported values use two decimal places, with aggregation and contrasts computed before rounding. The appendix provides all six metrics, seed variability, and a concise account of paired uncertainty.

\subsection{Main Results}
\label{sec:global_results}
Table~\ref{tab:global} compares each base retriever, \covercfg{}, and their fusion in both directions. A one-round validation sweep over 21 weights, spaced by 0.05, selected $\gamma=0.75$ by mean bidirectional macro Rsum3. At this fixed weight, two rounds achieve the highest bidirectional validation mean among the three refinement policies (Sec.~\ref{sec:camera_rounds}). The coefficient is shared across methods, text towers, datasets, directions, and camera seeds. Appendix~\ref{sec:supp_main_bidirectional} reports the complete results.

\begin{table}[!htbp]
\centering
\caption{\textbf{Bidirectional scene retrieval.} Each indented +\method{} row combines \method{} with the base retriever immediately above. Bold marks the better value within each pair.}
\label{tab:global}
\small\setlength{\tabcolsep}{3pt}
\renewcommand{\arraystretch}{1.01}
\begin{tabular}{@{}lrrr@{\hspace{7pt}}rrr@{\hspace{9pt}}rrr@{\hspace{7pt}}rrr@{}}
\toprule
& \multicolumn{6}{c}{ScanNet} & \multicolumn{6}{c}{3RScan} \\
\cmidrule(lr){2-7}\cmidrule(lr){8-13}
Method & \multicolumn{3}{c}{Text-to-scene} & \multicolumn{3}{c}{Scene-to-text} & \multicolumn{3}{c}{Text-to-scene} & \multicolumn{3}{c}{Scene-to-text} \\
\cmidrule(lr){2-4}\cmidrule(lr){5-7}\cmidrule(lr){8-10}\cmidrule(lr){11-13}
& R@1 & Rsum3 & MRR & R@1 & Rsum3 & MRR & R@1 & Rsum3 & MRR & R@1 & Rsum3 & MRR \\
\midrule
\covercfg{} & 11.80 & 90.95 & 0.23 & 6.87 & 73.54 & 0.17 & 8.28 & 73.38 & 0.19 & 6.49 & 69.57 & 0.17 \\
\midrule
\multicolumn{13}{@{}l}{\textbf{BiGRU}}\\
VSE$\infty$ & 0.00 & 5.50 & 0.02 & 0.69 & 7.22 & 0.03 & 0.67 & 12.08 & 0.04 & 0.67 & 12.08 & 0.04 \\
\quad +\method{} & \textbf{10.42} & \textbf{81.90} & \textbf{0.21} & \textbf{5.61} & \textbf{70.68} & \textbf{0.16} & \textbf{7.61} & \textbf{63.53} & \textbf{0.17} & \textbf{4.47} & \textbf{55.48} & \textbf{0.14} \\
ESA & 1.72 & 22.68 & 0.06 & 1.03 & 9.97 & 0.04 & 0.67 & 22.15 & 0.06 & 2.01 & 18.79 & 0.06 \\
\quad +\method{} & \textbf{12.03} & \textbf{94.27} & \textbf{0.23} & \textbf{7.67} & \textbf{75.03} & \textbf{0.18} & \textbf{7.61} & \textbf{78.08} & \textbf{0.18} & \textbf{7.61} & \textbf{73.38} & \textbf{0.18} \\
CRCL & 2.41 & 28.87 & 0.07 & 1.37 & 23.37 & 0.06 & 1.34 & 23.49 & 0.07 & 0.00 & 12.75 & 0.04 \\
\quad +\method{} & \textbf{12.49} & \textbf{95.99} & \textbf{0.24} & \textbf{8.25} & \textbf{79.04} & \textbf{0.19} & \textbf{5.59} & \textbf{69.57} & \textbf{0.17} & \textbf{6.71} & \textbf{65.32} & \textbf{0.16} \\
IDeal Source & 7.56 & 64.26 & 0.16 & 5.50 & 53.95 & 0.14 & 4.70 & 68.46 & 0.15 & 6.71 & 64.43 & 0.15 \\
\quad +\method{} & \textbf{12.26} & \textbf{97.37} & \textbf{0.25} & \textbf{9.28} & \textbf{84.19} & \textbf{0.20} & \textbf{11.63} & \textbf{92.39} & \textbf{0.23} & \textbf{7.38} & \textbf{81.21} & \textbf{0.20} \\
RoMa & 7.22 & 65.64 & 0.16 & 5.84 & 56.70 & 0.15 & 9.40 & 73.15 & 0.18 & 4.70 & 60.40 & 0.15 \\
\quad +\method{} & \textbf{12.94} & \textbf{101.95} & \textbf{0.26} & \textbf{9.62} & \textbf{95.42} & \textbf{0.22} & \textbf{10.96} & \textbf{91.28} & \textbf{0.23} & \textbf{8.95} & \textbf{80.54} & \textbf{0.20} \\
\midrule
\multicolumn{13}{@{}l}{\textbf{BERT}}\\
VSE$\infty$ & 2.75 & 34.36 & 0.09 & 2.75 & 30.58 & 0.08 & 2.68 & 37.58 & 0.10 & 4.03 & 38.26 & 0.10 \\
\quad +\method{} & \textbf{12.94} & \textbf{97.14} & \textbf{0.25} & \textbf{8.36} & \textbf{80.87} & \textbf{0.19} & \textbf{8.28} & \textbf{77.85} & \textbf{0.19} & \textbf{6.94} & \textbf{71.14} & \textbf{0.18} \\
ESA & 0.34 & 5.84 & 0.02 & 0.34 & 5.50 & 0.02 & 0.67 & 10.74 & 0.04 & 0.67 & 10.74 & 0.04 \\
\quad +\method{} & \textbf{10.42} & \textbf{83.62} & \textbf{0.21} & \textbf{8.02} & \textbf{69.76} & \textbf{0.17} & \textbf{8.05} & \textbf{70.92} & \textbf{0.17} & \textbf{6.71} & \textbf{64.88} & \textbf{0.16} \\
CRCL & 1.37 & 24.05 & 0.06 & 1.37 & 20.27 & 0.06 & 2.68 & 31.54 & 0.09 & 3.36 & 32.89 & 0.09 \\
\quad +\method{} & \textbf{11.91} & \textbf{95.42} & \textbf{0.24} & \textbf{8.71} & \textbf{79.15} & \textbf{0.19} & \textbf{7.38} & \textbf{75.62} & \textbf{0.18} & \textbf{8.05} & \textbf{76.73} & \textbf{0.18} \\
IDeal Source & 6.87 & 64.26 & 0.15 & 4.81 & 60.82 & 0.15 & 6.04 & 71.81 & 0.16 & 5.37 & 72.48 & 0.17 \\
\quad +\method{} & \textbf{12.83} & \textbf{100.00} & \textbf{0.25} & \textbf{9.05} & \textbf{89.12} & \textbf{0.21} & \textbf{11.86} & \textbf{95.97} & \textbf{0.24} & \textbf{9.40} & \textbf{86.58} & \textbf{0.21} \\
RoMa & 7.22 & 64.95 & 0.16 & 6.53 & 67.01 & 0.16 & 5.37 & 71.81 & 0.16 & 7.38 & 69.80 & 0.17 \\
\quad +\method{} & \textbf{12.14} & \textbf{101.60} & \textbf{0.25} & \textbf{11.68} & \textbf{91.98} & \textbf{0.23} & \textbf{11.19} & \textbf{91.28} & \textbf{0.23} & \textbf{9.17} & \textbf{85.23} & \textbf{0.21} \\
\midrule
\multicolumn{13}{@{}l}{\textbf{Model-specific text encoders}}\\
CLIP2CLIP & 2.75 & 27.84 & 0.08 & 2.06 & 24.74 & 0.07 & 1.34 & 22.15 & 0.06 & 2.01 & 16.78 & 0.06 \\
\quad +\method{} & \textbf{7.33} & \textbf{76.86} & \textbf{0.18} & \textbf{7.90} & \textbf{70.10} & \textbf{0.17} & \textbf{7.61} & \textbf{65.32} & \textbf{0.16} & \textbf{6.26} & \textbf{60.18} & \textbf{0.16} \\
CrossOver & 6.19 & 68.04 & 0.16 & 5.84 & 62.89 & 0.15 & 7.38 & 67.11 & 0.17 & 8.72 & 70.47 & 0.18 \\
\quad +\method{} & \textbf{13.63} & \textbf{103.32} & \textbf{0.26} & \textbf{10.31} & \textbf{92.67} & \textbf{0.22} & \textbf{7.61} & \textbf{85.91} & \textbf{0.20} & \textbf{9.62} & \textbf{84.56} & \textbf{0.21} \\
\bottomrule
\end{tabular}
\end{table}

\begin{figure}[!htbp]
\centering
\includegraphics[width=.65\linewidth]{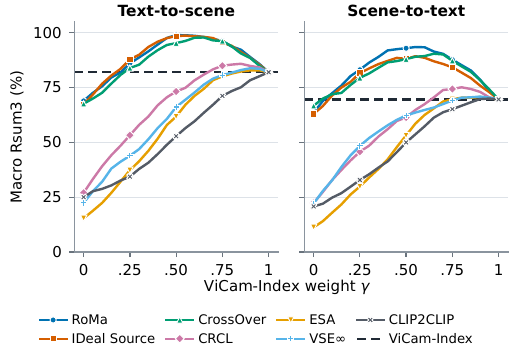}
\caption{\textbf{Family-wise fusion sensitivity.} The one-round 21-point validation grid varies the \covercfg{} weight in $F_\gamma=(1-\gamma)Z(G)+\gamma Z(I)$. Curves average datasets, text towers, and three camera seeds. Dashed lines mark standalone \covercfg{} (81.92/69.54). Broad intervals above them indicate complementarity; crossings near $\gamma=1$ are visual-dominated.}
\label{fig:family_sensitivity}
\end{figure}

Fusion improves Rsum3 over its corresponding base in all 48 dataset, direction, and text-encoder configurations. The equal-family base, visual-only, and fused means are 41.91, 82.16, and 86.38 for text-to-scene and 38.49, 71.56, and 77.53 for scene-to-text. Figure~\ref{fig:family_sensitivity} exposes the full family-wise response. Because base-to-fusion gains also grow when the base is weak, they do not alone establish complementarity.

ESA, CRCL, and VSE$\infty$ are image--text retrieval architectures retrained here on frozen DGCNN point-patch descriptors~\cite{zhu2023esa,qin2023crcl,chen2021vseinfty}. Their base Rsum3 ranges are only 15.35--26.99 (text-to-scene) and 11.25--22.32 (scene-to-text); their visual-heavy optima ($\gamma=0.80$--$0.95$) lie just 0.88--3.85 and 1.11--5.53 points above \covercfg{}, respectively. These rows mainly demonstrate a strong drop-in visual retriever. CLIP2CLIP is instead a two-CLIP control: frozen OpenAI CLIP on \initcfg{} views is combined with EVA-CLIP on refined views. Its 0.00/0.39-point text-to-scene/scene-to-text gains indicate redundant visual evidence, not point--text/visual complementarity. RoMa, IDeal Source, and CrossOver peak at $\gamma=0.50$--$0.65$, add 15.83--16.94/19.70--23.87 points in the two directions, and remain above \covercfg{} over broad intervals; these families provide the main complementarity evidence.

\FloatBarrier

\subsection{Ablation Studies}
The following summaries isolate the main effect of each design choice. Full protocols, bidirectional tables, seed statistics, and geometric diagnostics are provided in the appendix to keep the main text focused on the findings.

\paragraph{The impact of camera initialization.}
\label{sec:initialization_ablation}
\label{sec:selection_results}
With candidate pools, rendering, encoding, and dataset-specific budgets fixed, \initcfg{} achieves macro text-to-scene Rsum3 of 83.40, compared with 55.82 for Uniform and 54.26 for five-seed Random. These controls evaluate the initial views before learned refinement (Appendix~\ref{sec:dataset_budget_selection}).

\paragraph{The impact of observation budget.}
\label{sec:budget}
\begin{table}[!htbp]
\centering\small
\setlength{\tabcolsep}{5pt}
\caption{\textbf{Observation-budget ablation.} \initcfg{} standalone text-to-scene macro Rsum3 across both datasets.}
\label{tab:observation_budget}
\begin{tabular}{@{}lccccc@{}}
\toprule
Views $K$ & 6 & 12 & 18 & 24 & 30 \\
\midrule
Rsum3 & 81.00 & \textbf{81.39} & 81.38 & 78.00 & 78.07 \\
\bottomrule
\end{tabular}
\end{table}
Table~\ref{tab:observation_budget} shows that retrieval peaks at $K=12$ and does not improve monotonically with more views. From 6 to 30 views, mean preparation time increases from 12.08 to 14.61 seconds per scene and descriptor storage from 10.31 to 51.21 MiB. Appendix~\ref{sec:budget_cost} reports the per-dataset results and measurement conditions.

\paragraph{The impact of camera supervision.}
\label{sec:camera_supervision}
Among five coefficients, $\lambda=0.50$ gives the highest mean ScanNet text-to-scene Rsum3, 93.13, compared with 86.94 for fixed \initcfg{}. The best observed coefficient varies across datasets and retrieval directions (Appendix~\ref{sec:supp_camera_lambda_sweep}).

\paragraph{The impact of pose degrees of freedom.}
\label{sec:camera_pose_dof}
At $\lambda=0.50$, XYZ/yaw/pitch refinement gives the highest mean text-to-scene Rsum3 on ScanNet (93.13). Individual angular contributions remain uncertain across the three matched seeds (Appendix~\ref{sec:supp_camera_pose_dof}).

\paragraph{The impact of geometric acceptance.}
\label{sec:camera_acceptance}
Adding the gain check to geometric validity increases the final geometric objective on both datasets. Text-to-scene Rsum3 changes by $+2.29$ points on ScanNet and $-1.12$ on 3RScan (Appendix~\ref{sec:supp_camera_acceptance}).

\begin{figure}[!htbp]
\centering
\includegraphics[width=.70\linewidth]{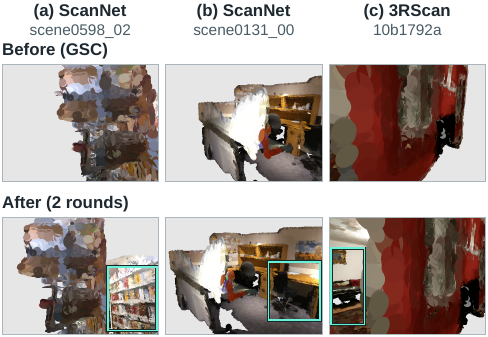}
\caption{\textbf{Scene detail revealed by camera refinement.}
Each column shows Surfel renderings of one validation scene from the same camera slot at its initial \initcfg{} pose (top) and after two accepted camera actions (bottom).
Cyan boxes highlight (a) newly visible shelving, (b) more of the desk and office chair, and (c) wall art and a rug beyond the foreground occluder.}
\label{fig:camera_refinement_examples}
\end{figure}

\paragraph{The impact of refinement rounds.}
\label{sec:camera_rounds}
Relative to two rounds, adaptive refinement adds 2.92 points to macro visual text-to-scene Rsum3 (Appendix~\ref{sec:supp_camera_rounds}).
Figure~\ref{fig:camera_refinement_examples} shows the additional scene detail visible after two accepted updates.

\FloatBarrier

\paragraph{The impact of the vision--language backbone.}
\label{sec:clip_backbone_ablation}
EVA remains strongest in both retrieval directions, with OpenCLIP bigG/14 the closest substitute (Table~\ref{tab:clip_backbone_ablation}). Appendix~\ref{sec:supp_clip_backbone} reports the visual-only results and per-dataset metrics.
\begin{table}[!htbp]
\centering
\setlength{\tabcolsep}{5pt}
\caption{\textbf{Vision--language backbone ablation.} Fused macro Rsum3 at $\gamma=0.75$.}
\label{tab:clip_backbone_ablation}
\begin{tabular}{@{}lcc@{}}
\toprule
Backbone & Text$\to$scene & Scene$\to$text \\
\midrule
EVA02-E-14-plus~\cite{evaclip} & \textbf{86.38} & \textbf{77.53} \\
OpenCLIP bigG/14~\cite{cherti2023reproducible} & 68.92 & 65.00 \\
SigLIP 2 Base/16~\cite{tschannen2025siglip2} & 62.39 & 41.81 \\
\bottomrule
\end{tabular}
\end{table}

\section{Conclusion and Limitations}
\label{sec:conclusion}
\method{} provides a geometry-supervised virtual camera pipeline to build highly effective, reusable visual indices for bidirectional text--3D scene retrieval. By combining Greedy Surface Coverage with learned pose refinement, it dynamically captures discriminative views from colored point clouds without manual annotations. Crucially, utilizing surfel rendering and frozen image--text models completely bypasses the computational bottleneck of per-scene neural fitting. Under a compact 12-view budget, our fused approach achieves 86.38 and 77.53 macro Rsum3 for text-to-scene and scene-to-text retrieval on ScanNet and 3RScan. Furthermore, evaluations across seven retriever families yield a critical insight: while ensembling pure 2D models produces redundant visual evidence, seamlessly fusing our geometry-aware 2D index with 3D-native retrievers unlocks broad complementary gains.

Despite its plug-and-play efficiency, our approach has limitations. The evaluation assumes metric, Z-up indoor reconstructions, and surfel quality heavily relies on initial point density and estimated normals. Furthermore, because the reported validation galleries informed our development, checkpoint, and hyperparameter selection, independent locked evaluation is required to fully assess generalization. Addressing these constraints and exploring alternative camera initializers remain valuable directions for future work to further bridge the modality gap in cross-modal 3D understanding.

\FloatBarrier
{\small
  \setlength{\bibsep}{2pt plus 0.5pt}
  \bibliographystyle{ieeenat_fullname}
  \bibliography{references}

@inproceedings{chen2020scanrefer,
  author    = {Chen, Dave Zhenyu and Chang, Angel X. and Nie{\ss}ner, Matthias},
  title     = {{ScanRefer}: {3D} Object Localization in {RGB-D} Scans Using Natural Language},
  booktitle = {Computer Vision -- ECCV},
  pages     = {202--221},
  year      = {2020}
}

@inproceedings{achlioptas2020referit3d,
  author    = {Achlioptas, Panos and Abdelreheem, Ahmed and Xia, Fei and
               Elhoseiny, Mohamed and Guibas, Leonidas J.},
  title     = {{ReferIt3D}: Neural Listeners for Fine-Grained {3D} Object
               Identification in Real-World Scenes},
  booktitle = {Computer Vision -- ECCV},
  pages     = {422--440},
  year      = {2020}
}

@article{roma,
  title={{PointCloud-Text Matching: Benchmark Dataset and Baseline}},
  author={Feng, Yanglin and Qin, Yang and Peng, Dezhong and Zhu, Hongyuan and Peng, Xi and Hu, Peng},
  journal={IEEE Transactions on Multimedia},
  year={2025},
  publisher={IEEE},
  url={https://arxiv.org/abs/2403.19386}
}

@article{evaclip,
 title={{EVA-CLIP: Improved Training Techniques for CLIP at Scale}},
 author={Sun, Quan and Fang, Yuxin and Wu, Ledell and Wang, Xinlong and Cao, Yue},
 journal={arXiv preprint arXiv:2303.15389}, year={2023},
 url={https://arxiv.org/abs/2303.15389}
}

@inproceedings{cherti2023reproducible,
  author={Cherti, Mehdi and Beaumont, Romain and Wightman, Ross and Wortsman, Mitchell and
          Ilharco, Gabriel and Gordon, Cade and Schuhmann, Christoph and Schmidt, Ludwig and Jitsev, Jenia},
  title={Reproducible Scaling Laws for Contrastive Language-Image Learning},
  booktitle={Proceedings of the IEEE/CVF Conference on Computer Vision and Pattern Recognition},
  pages={2818--2829},
  year={2023},
  url={https://openaccess.thecvf.com/content/CVPR2023/papers/Cherti_Reproducible_Scaling_Laws_for_Contrastive_Language-Image_Learning_CVPR_2023_paper.pdf}
}

@article{tschannen2025siglip2,
  author={Tschannen, Michael and Gritsenko, Alexey and Wang, Xiao and Naeem, Muhammad Ferjad and
          Alabdulmohsin, Ibrahim and Parthasarathy, Nikhil and Evans, Talfan and Beyer, Lucas and
          Xia, Ye and Mustafa, Basil and H{\'e}naff, Olivier and Harmsen, Jeremiah and
          Steiner, Andreas and Zhai, Xiaohua},
  title={{SigLIP 2}: Multilingual Vision-Language Encoders with Improved Semantic Understanding,
         Localization, and Dense Features},
  journal={arXiv preprint arXiv:2502.14786},
  year={2025},
  url={https://arxiv.org/abs/2502.14786}
}

@inproceedings{feng2025interactive,
  author={Feng, Yanglin and Li, Yongxiang and Sun, Yuan and Qin, Yang and Peng, Dezhong and Hu, Peng},
  title={{Interactive Cross-modal Learning for Text-3D Scene Retrieval}},
  booktitle={{Advances in Neural Information Processing Systems}},
  volume={38, Main Conference},
  pages={97972--97992},
  publisher={Curran Associates, Inc.},
  year={2025},
  doi={10.52202/085713-3273},
  url={https://proceedings.neurips.cc/paper_files/paper/2025/file/8ce0b141fb61bb847898d0b6258f9acf-Paper-Conference.pdf}
}

@inproceedings{crossover,
  author={Sarkar, Sayan Deb and Miksik, Ondrej and Pollefeys, Marc and Barath, Daniel and Armeni, Iro},
  title={{CrossOver: 3D Scene Cross-Modal Alignment}},
  booktitle={{Proceedings of the IEEE/CVF Conference on Computer Vision and Pattern Recognition}},
  month={June},
  year={2025},
  pages={8985--8994},
  url={https://openaccess.thecvf.com/content/CVPR2025/html/Sarkar_CrossOver_3D_Scene_Cross-Modal_Alignment_CVPR_2025_paper.html}
}

@inproceedings{retri3d,
  title={{Retri3D: 3D Neural Graphics Representation Retrieval}},
  author={Guan, Yushi and Kwan, Daniel and Dandurand, Jean Sebastien and Yan, Xi and Liang, Ruofan and Zhang, Yuxuan and Jain, Nilesh and Ahuja, Nilesh and Panneer, Selvakumar and Vijaykumar, Nandita},
  booktitle={{International Conference on Learning Representations}},
  year={2025},
  url={https://openreview.net/forum?id=q3EbOXb4y1}
}

@inproceedings{whereami,
  title={{{``Where am I?''} Scene Retrieval with Language}},
  author={Chen, Jiaqi and Barath, Daniel and Armeni, Iro and Pollefeys, Marc and Blum, Hermann},
  booktitle={{European Conference on Computer Vision}},
  year={2024},
  url={https://www.ecva.net/papers/eccv_2024/papers_ECCV/papers/05418.pdf}
}

@inproceedings{langloc,
  title={{LangLoc: Tell Me What You See}},
  author={Panwar, Shaurya Kishore and Zendehdel Nobari, Roham and Lau, Shirley Feng Yi and Shaik, Abu Bakr Rahman and G{\"u}nther, Manuel and Pollefeys, Marc and Barath, Daniel},
  booktitle={{European Conference on Computer Vision}},
  year={2026},
  url={https://arxiv.org/abs/2607.05077}
}

@InProceedings{clip,
 title = {{Learning Transferable Visual Models From Natural Language Supervision}},
 author = {Radford, Alec and Kim, Jong Wook and Hallacy, Chris and Ramesh, Aditya and Goh, Gabriel and Agarwal, Sandhini and Sastry, Girish and Askell, Amanda and Mishkin, Pamela and Clark, Jack and Krueger, Gretchen and Sutskever, Ilya},
 booktitle = {{Proceedings of the 38th International Conference on Machine Learning}},
 pages = {8748--8763}, year = {2021},
 editor = {Meila, Marina and Zhang, Tong}, volume = {139},
 series = {Proceedings of Machine Learning Research},
 month = {18--24 Jul}, publisher = {PMLR},
 url = {https://proceedings.mlr.press/v139/radford21a.html}
}

@inproceedings{openins3d,
 title={{OpenIns3D: Snap and Lookup for 3D Open-vocabulary Instance Segmentation}},
 author={Zhening Huang and Xiaoyang Wu and Xi Chen and Hengshuang Zhao and Lei Zhu and Joan Lasenby},
 booktitle={{European Conference on Computer Vision}},
 year={2024}
}

@article{uniscene3d,
 title={{RGB-Pointmap Pretraining for Unified 3D Scene Understanding}},
 author={Mao, Ye and Luo, Weixun and Huang, Ranran and Jing, Junpeng and Mikolajczyk, Krystian},
 journal={arXiv preprint arXiv:2604.02546}, year={2026},
 url={https://arxiv.org/abs/2604.02546v3}
}

@InProceedings{qbnorm,
    author    = {Bogolin, Simion-Vlad and Croitoru, Ioana and Jin, Hailin and Liu, Yang and Albanie, Samuel},
    title     = {Cross Modal Retrieval With Querybank Normalisation},
    booktitle = {Proceedings of the IEEE/CVF Conference on Computer Vision and Pattern Recognition (CVPR)},
    month     = {June},
    year      = {2022},
    pages     = {5194-5205}
}

@inproceedings{dbnorm,
 title={{Balance Act: Mitigating Hubness in Cross-Modal Retrieval with Query and Gallery Banks}},
 author={Wang, Yimu and Jian, Xiangru and Xue, Bo},
 booktitle={Proceedings of the 2023 Conference on Empirical Methods in Natural Language Processing},
 pages={10542--10567}, year={2023}, publisher={Association for Computational Linguistics},
 doi={10.18653/v1/2023.emnlp-main.652}, url={https://aclanthology.org/2023.emnlp-main.652/}
}

@inproceedings{sun2021viewselection,
  author = {Sun, Yifan and Huang, Qixing and Hsiao, Dun-Yu and Guan, Li and Hua, Gang},
  title = {{Learning View Selection for 3D Scenes}},
  booktitle = {Proceedings of the IEEE/CVF Conference on Computer Vision and Pattern Recognition},
  year = {2021},
  pages = {14464--14473},
  url = {https://openaccess.thecvf.com/content/CVPR2021/html/Sun_Learning_View_Selection_for_3D_Scenes_CVPR_2021_paper.html}
}

@inproceedings{hamdi2021mvtn,
  author = {Hamdi, Abdullah and Giancola, Silvio and Ghanem, Bernard},
  title = {{MVTN: Multi-View Transformation Network for 3D Shape Recognition}},
  booktitle = {Proceedings of the IEEE/CVF International Conference on Computer Vision},
  year = {2021},
  pages = {1--11},
  url = {https://openaccess.thecvf.com/content/ICCV2021/html/Hamdi_MVTN_Multi-View_Transformation_Network_for_3D_Shape_Recognition_ICCV_2021_paper.html}
}

@inproceedings{markov2012normalization,
  author = {Markov, Ilya and Arampatzis, Avi and Crestani, Fabio},
  title = {{Unsupervised Linear Score Normalization Revisited}},
  booktitle = {Proceedings of the 35th International ACM SIGIR Conference on Research and Development in Information Retrieval},
  year = {2012},
  pages = {1161--1162},
  doi = {10.1145/2348283.2348519},
  url = {https://doi.org/10.1145/2348283.2348519}
}

@inproceedings{dai2017scannet,
  author={Angela Dai and Angel X. Chang and Manolis Savva and Maciej Halber and Thomas Funkhouser and Matthias Niessner},
  title={{ScanNet}: Richly-Annotated {3D} Reconstructions of Indoor Scenes},
  booktitle={Proceedings of the IEEE Conference on Computer Vision and Pattern Recognition},
  year={2017}, pages={2432--2443},
  url={https://openaccess.thecvf.com/content_cvpr_2017/html/Dai_ScanNet_Richly-Annotated_3D_CVPR_2017_paper.html}
}

@inproceedings{wald2019rio,
  author={Johanna Wald and Armen Avetisyan and Nassir Navab and Federico Tombari and Matthias Niessner},
  title={{RIO}: {3D} Object Instance Re-Localization in Changing Indoor Environments},
  booktitle={Proceedings of the IEEE/CVF International Conference on Computer Vision},
  year={2019}, pages={7657--7666},
  url={https://waldjohannau.github.io/RIO/}
}

@inproceedings{jia2024sceneverse,
  author={Baoxiong Jia and Yixin Chen and Huangyue Yu and Yan Wang and Xuesong Niu and Tengyu Liu and Qing Li and Siyuan Huang},
  title={{SceneVerse}: Scaling {3D} Vision-Language Learning for Grounded Scene Understanding},
  booktitle={European Conference on Computer Vision}, year={2024},
  url={https://github.com/scene-verse/SceneVerse}
}

@inproceedings{louis2025know,
  author={Antoine Louis and Gijs van Dijck and Gerasimos Spanakis},
  title={Know When to Fuse: Investigating Non-English Hybrid Retrieval in the Legal Domain},
  booktitle={Proceedings of the 31st International Conference on Computational Linguistics},
  year={2025}, pages={4293--4312},
  publisher={Association for Computational Linguistics},
  url={https://aclanthology.org/2025.coling-main.290/}
}

@article{zhu2023esa,
  author={Hongguang Zhu and Chunjie Zhang and Yunchao Wei and Shujuan Huang and Yao Zhao},
  title={{ESA}: External Space Attention Aggregation for Image-Text Retrieval},
  journal={IEEE Transactions on Circuits and Systems for Video Technology},
  year={2023}, url={https://github.com/KevinLight831/ESA}
}

@inproceedings{qin2023crcl,
  author={Yang Qin and Yuan Sun and Dezhong Peng and Joey Tianyi Zhou and Xi Peng and Peng Hu},
  title={Cross-modal Active Complementary Learning with Self-refining Correspondence},
  booktitle={Advances in Neural Information Processing Systems}, volume={36},
  year={2023},
  url={https://papers.nips.cc/paper_files/paper/2023/hash/4e580cdd54fe38ca9a5b8ea6fe99bb44-Abstract-Conference.html}
}

@inproceedings{chen2021vseinfty,
  author={Jiacheng Chen and Hexiang Hu and Hao Wu and Yuning Jiang and Changhu Wang},
  title={Learning the Best Pooling Strategy for Visual Semantic Embedding},
  booktitle={Proceedings of the IEEE/CVF Conference on Computer Vision and Pattern Recognition},
  year={2021}, url={https://github.com/woodfrog/vse_infty}
}

@article{nemhauser1978submodular,
  author={Nemhauser, George L. and Wolsey, Laurence A. and Fisher, Marshall L.},
  title={{An Analysis of Approximations for Maximizing Submodular Set Functions---I}},
  journal={Mathematical Programming},
  volume={14}, pages={265--294}, year={1978},
  url={https://thibaut.horel.org/submodularity/papers/nemhauser1978.pdf}
}

@inproceedings{zheng2025video3d,
  author={Zheng, Duo and Huang, Shijia and Wang, Liwei},
  title={{Video-3D LLM: Learning Position-Aware Video Representation for 3D Scene Understanding}},
  booktitle={Proceedings of the IEEE/CVF Conference on Computer Vision and Pattern Recognition},
  year={2025},
  url={https://openaccess.thecvf.com/content/CVPR2025/papers/Zheng_Video-3D_LLM_Learning_Position-Aware_Video_Representation_for_3D_Scene_Understanding_CVPR_2025_paper.pdf}
}

@inproceedings{mao2026poma3d,
  author={Mao, Ye and Luo, Weixun and Huang, Ranran and Jing, Junpeng and Mikolajczyk, Krystian},
  title={{POMA-3D: The Point Map Way to 3D Scene Understanding}},
  booktitle={Proceedings of the IEEE/CVF Conference on Computer Vision and Pattern Recognition (CVPR) Findings},
  year={2026},
  pages={7282--7292},
  url={https://openaccess.thecvf.com/content/CVPR2026F/html/Mao_POMA-3D_The_Point_Map_Way_to_3D_Scene_Understanding_CVPRF_2026_paper.html}
}

@inproceedings{chen2026converge,
  author={Chen, Timothy and Dai, Adam and Adang, Maximilian and Gao, Grace and Schwager, Mac},
  title={{Coverage Optimization for Camera View Selection}},
  booktitle={Proceedings of the IEEE/CVF Conference on Computer Vision and Pattern Recognition},
  year={2026},
  pages={19443--19451},
  url={https://openaccess.thecvf.com/content/CVPR2026/html/Chen_Coverage_Optimization_for_Camera_View_Selection_CVPR_2026_paper.html}
}
}

\clearpage
\appendix
\pdfbookmark[0]{Appendix}{appendix}
\part*{Appendix}
\section{Evaluation Protocol}
This supplement details the CrossOver-split ScanNet/3RScan experiments. The paired validation galleries contain 291 and 149 scenes, respectively. Both retrieval directions share raw scores, with candidate-axis normalization and fusion recomputed for each direction. Macro results weight datasets equally; seven-family means first average BERT and BiGRU within each applicable family, then weight families equally. CLIP2CLIP and CrossOver each retain their native text encoder. \basecfg{} denotes an existing retriever, \covercfg{} the standalone visual retriever, and each named retriever+\method{} entry their score-level combination.

The virtual observation branch constructs its index directly from metric, Z-up colored point clouds, without acquisition image--pose pairs, external pointmaps, or per-scene neural fitting. ScanNet vertex RGB and decoded 3RScan OBJ/MTL textures supply the colors. Camera placement and visibility use XYZ.

\paragraph{Camera budgets and refinement rounds.} The main comparison and backbone substitution use $K=12$ on both datasets, two refinement rounds, $\lambda=0.50$, and visual fusion weight $\gamma=0.75$. The round ablation reuses these checkpoints for one, two, or adaptive rounds. The initialization, supervision, pose, and acceptance studies use ScanNet $K=12$ and 3RScan $K=6$: initialization evaluates fixed poses, and the three learned-policy ablations use one refinement round. Their camera models are fitted separately from the uniform-$K=12$ models, using matched seeds and training scenes. The observation-budget study evaluates fixed \initcfg{} poses at $K\in\{6,12,18,24,30\}$.

\subsection{Metrics and Uncertainty}
\label{sec:uncertainty}
We report R@1, R@5, R@10, R@20, Rsum3 (R@1+R@5+R@10), and MRR. Recall and Rsum3 are in percentage points; MRR uses a 0--1 scale. Values are displayed to two decimal places, with aggregation and contrasts computed before rounding. Learned results average three camera seeds; $\pm$ denotes their sample standard deviation where shown. Random uses five view-bank seeds. Main, round, and backbone tables report means.

Paired bootstrap analyses retain the full galleries and resample physical-source families (138 on ScanNet and 46 on 3RScan). Initialization controls use 20,000 draws; learned-camera studies use 5,000 draws and also resample matched camera seeds. Pointwise 95\% intervals quantify variation over these resampled units conditional on the saved galleries, checkpoints, and view banks. Base-training, bank-regeneration, and validation-selection variability fall outside this scope.

\FloatBarrier
\section{Complete Bidirectional Main Results}
\label{sec:supp_main_bidirectional}
Tables~\ref{tab:supp_main_referral_to_point} and~\ref{tab:supp_main_point_to_referral}
report all six metrics for two refinement rounds at $K=12$ and $\lambda=0.50$. Each camera seed
has its own rankings, and the tables average their metrics. All seeds share the original base scores.
Macro Rsum3 (mean $\pm$ sample SD) is $82.16\pm1.44$ for the visual index and
$86.38\pm1.17$ after fusion in text-to-scene retrieval; the scene-to-text values
are $71.56\pm0.25$ and $77.53\pm0.54$. Each seed is aggregated across text
encoders, families, and datasets before computing the standard deviation.

The one-round validation grid selected $\gamma=0.75$ from 21 candidates spaced by 0.05 using bidirectional
macro Rsum3, then MRR, R@1, and the smaller $\gamma$ for ties. We use the fixed $\gamma=0.75$ across the round ablation.
At this weight, the one-round, two-round, and adaptive policies reach bidirectional macro Rsum3 of 81.53, 81.95, and 81.85 on the
validation galleries. The two-round result is the main protocol. Appendix~\ref{sec:supp_camera_rounds} gives the refinement-round comparison.
\begin{table}[!htbp]
\centering\footnotesize\setlength{\tabcolsep}{3pt}
\caption{\textbf{Complete text-to-scene results at $\gamma=0.75$.} The main protocol uses two refinement rounds at $K=12$ and $\lambda=0.50$ on both datasets. \covercfg{} and combined rows give three-seed mean metrics. Bold compares the base/combined pair at displayed precision.}
\label{tab:supp_main_referral_to_point}
\begin{tabular}{@{}lrrrrrr|rrrrrr@{}}
\toprule
Method & \multicolumn{6}{c|}{ScanNet} & \multicolumn{6}{c}{3RScan} \\
\cmidrule(lr){2-7}\cmidrule(lr){8-13}
& R@1 & R@5 & R@10 & R@20 & Rsum3 & MRR & R@1 & R@5 & R@10 & R@20 & Rsum3 & MRR \\
\midrule
\covercfg{} & 11.80 & 32.19 & 46.96 & 63.34 & 90.95 & 0.23 & 8.28 & 26.85 & 38.26 & 58.17 & 73.38 & 0.19 \\
\midrule
\multicolumn{13}{@{}l}{\textbf{BiGRU}}\\
VSE$\infty$ & 0.00 & 1.72 & 3.78 & 6.53 & 5.50 & 0.02 & 0.67 & 4.03 & 7.38 & 13.42 & 12.08 & 0.04 \\
\quad \combinedcfg{VSE$\infty$} & \textbf{10.42} & \textbf{29.21} & \textbf{42.27} & \textbf{59.45} & \textbf{81.90} & \textbf{0.21} & \textbf{7.61} & \textbf{22.82} & \textbf{33.11} & \textbf{52.13} & \textbf{63.53} & \textbf{0.17} \\
ESA & 1.72 & 8.25 & 12.71 & 20.62 & 22.68 & 0.06 & 0.67 & 7.38 & 14.09 & 25.50 & 22.15 & 0.06 \\
\quad \combinedcfg{ESA} & \textbf{12.03} & \textbf{34.82} & \textbf{47.42} & \textbf{64.95} & \textbf{94.27} & \textbf{0.23} & \textbf{7.61} & \textbf{26.85} & \textbf{43.62} & \textbf{58.39} & \textbf{78.08} & \textbf{0.18} \\
CRCL & 2.41 & 8.93 & 17.53 & 26.12 & 28.87 & 0.07 & 1.34 & 7.38 & 14.77 & 26.85 & 23.49 & 0.07 \\
\quad \combinedcfg{CRCL} & \textbf{12.49} & \textbf{34.71} & \textbf{48.80} & \textbf{67.47} & \textbf{95.99} & \textbf{0.24} & \textbf{5.59} & \textbf{22.82} & \textbf{41.16} & \textbf{59.28} & \textbf{69.57} & \textbf{0.17} \\
IDeal Source & 7.56 & 23.02 & 33.68 & 45.70 & 64.26 & 0.16 & 4.70 & 24.16 & 39.60 & 53.69 & 68.46 & 0.15 \\
\quad \combinedcfg{IDeal Source} & \textbf{12.26} & \textbf{35.74} & \textbf{49.37} & \textbf{71.82} & \textbf{97.37} & \textbf{0.25} & \textbf{11.63} & \textbf{32.89} & \textbf{47.87} & \textbf{66.44} & \textbf{92.39} & \textbf{0.23} \\
RoMa & 7.22 & 22.68 & 35.74 & 49.14 & 65.64 & 0.16 & 9.40 & 26.85 & 36.91 & 50.34 & 73.15 & 0.18 \\
\quad \combinedcfg{RoMa} & \textbf{12.94} & \textbf{35.51} & \textbf{53.49} & \textbf{71.71} & \textbf{101.95} & \textbf{0.26} & \textbf{10.96} & \textbf{34.23} & \textbf{46.09} & \textbf{60.85} & \textbf{91.28} & \textbf{0.23} \\
\midrule
\multicolumn{13}{@{}l}{\textbf{BERT}}\\
VSE$\infty$ & 2.75 & 11.34 & 20.27 & 38.49 & 34.36 & 0.09 & 2.68 & 12.75 & 22.15 & 37.58 & 37.58 & 0.10 \\
\quad \combinedcfg{VSE$\infty$} & \textbf{12.94} & \textbf{35.40} & \textbf{48.80} & \textbf{69.30} & \textbf{97.14} & \textbf{0.25} & \textbf{8.28} & \textbf{27.96} & \textbf{41.61} & \textbf{60.63} & \textbf{77.85} & \textbf{0.19} \\
ESA & 0.34 & 1.72 & 3.78 & 6.87 & 5.84 & 0.02 & 0.67 & 3.36 & 6.71 & 14.09 & 10.74 & 0.04 \\
\quad \combinedcfg{ESA} & \textbf{10.42} & \textbf{30.24} & \textbf{42.96} & \textbf{58.88} & \textbf{83.62} & \textbf{0.21} & \textbf{8.05} & \textbf{25.28} & \textbf{37.58} & \textbf{52.13} & \textbf{70.92} & \textbf{0.17} \\
CRCL & 1.37 & 7.90 & 14.78 & 25.43 & 24.05 & 0.06 & 2.68 & 11.41 & 17.45 & 33.56 & 31.54 & 0.09 \\
\quad \combinedcfg{CRCL} & \textbf{11.91} & \textbf{36.08} & \textbf{47.42} & \textbf{65.64} & \textbf{95.42} & \textbf{0.24} & \textbf{7.38} & \textbf{27.74} & \textbf{40.49} & \textbf{58.61} & \textbf{75.62} & \textbf{0.18} \\
IDeal Source & 6.87 & 20.62 & 36.77 & 50.17 & 64.26 & 0.15 & 6.04 & 23.49 & 42.28 & 58.39 & 71.81 & 0.16 \\
\quad \combinedcfg{IDeal Source} & \textbf{12.83} & \textbf{35.97} & \textbf{51.20} & \textbf{69.99} & \textbf{100.00} & \textbf{0.25} & \textbf{11.86} & \textbf{35.79} & \textbf{48.32} & \textbf{63.76} & \textbf{95.97} & \textbf{0.24} \\
RoMa & 7.22 & 21.31 & 36.43 & 48.45 & 64.95 & 0.16 & 5.37 & 26.85 & 39.60 & 52.35 & 71.81 & 0.16 \\
\quad \combinedcfg{RoMa} & \textbf{12.14} & \textbf{37.00} & \textbf{52.46} & \textbf{71.59} & \textbf{101.60} & \textbf{0.25} & \textbf{11.19} & \textbf{33.56} & \textbf{46.53} & \textbf{66.00} & \textbf{91.28} & \textbf{0.23} \\
\midrule
\multicolumn{13}{@{}l}{\textbf{Model-specific text encoders}}\\
CLIP2CLIP & 2.75 & 9.62 & 15.46 & 22.34 & 27.84 & 0.08 & 1.34 & 8.72 & 12.08 & 19.46 & 22.15 & 0.06 \\
\quad \combinedcfg{CLIP2CLIP} & \textbf{7.33} & \textbf{27.03} & \textbf{42.50} & \textbf{59.34} & \textbf{76.86} & \textbf{0.18} & \textbf{7.61} & \textbf{23.71} & \textbf{34.00} & \textbf{52.13} & \textbf{65.32} & \textbf{0.16} \\
CrossOver & 6.19 & 25.43 & 36.43 & 50.86 & 68.04 & 0.16 & 7.38 & 24.16 & 35.57 & 53.69 & 67.11 & 0.17 \\
\quad \combinedcfg{CrossOver} & \textbf{13.63} & \textbf{37.11} & \textbf{52.58} & \textbf{71.71} & \textbf{103.32} & \textbf{0.26} & \textbf{7.61} & \textbf{29.31} & \textbf{48.99} & \textbf{66.22} & \textbf{85.91} & \textbf{0.20} \\
\bottomrule
\end{tabular}
\end{table}

\begin{table}[!htbp]
\centering\footnotesize\setlength{\tabcolsep}{3pt}
\caption{\textbf{Complete scene-to-text results at $\gamma=0.75$.} The main protocol uses two refinement rounds at $K=12$ and $\lambda=0.50$ on both datasets. \covercfg{} and combined rows give three-seed mean metrics. Bold compares the base/combined pair at displayed precision.}
\label{tab:supp_main_point_to_referral}
\begin{tabular}{@{}lrrrrrr|rrrrrr@{}}
\toprule
Method & \multicolumn{6}{c|}{ScanNet} & \multicolumn{6}{c}{3RScan} \\
\cmidrule(lr){2-7}\cmidrule(lr){8-13}
& R@1 & R@5 & R@10 & R@20 & Rsum3 & MRR & R@1 & R@5 & R@10 & R@20 & Rsum3 & MRR \\
\midrule
\covercfg{} & 6.87 & 26.69 & 39.98 & 57.39 & 73.54 & 0.17 & 6.49 & 27.07 & 36.02 & 50.56 & 69.57 & 0.17 \\
\midrule
\multicolumn{13}{@{}l}{\textbf{BiGRU}}\\
VSE$\infty$ & 0.69 & 2.75 & 3.78 & 7.22 & 7.22 & 0.03 & 0.67 & 4.03 & 7.38 & 14.09 & 12.08 & 0.04 \\
\quad \combinedcfg{VSE$\infty$} & \textbf{5.61} & \textbf{25.54} & \textbf{39.52} & \textbf{56.13} & \textbf{70.68} & \textbf{0.16} & \textbf{4.47} & \textbf{19.91} & \textbf{31.10} & \textbf{49.22} & \textbf{55.48} & \textbf{0.14} \\
ESA & 1.03 & 3.09 & 5.84 & 15.12 & 9.97 & 0.04 & 2.01 & 6.71 & 10.07 & 22.15 & 18.79 & 0.06 \\
\quad \combinedcfg{ESA} & \textbf{7.67} & \textbf{26.12} & \textbf{41.24} & \textbf{58.88} & \textbf{75.03} & \textbf{0.18} & \textbf{7.61} & \textbf{28.19} & \textbf{37.58} & \textbf{53.91} & \textbf{73.38} & \textbf{0.18} \\
CRCL & 1.37 & 6.87 & 15.12 & 26.12 & 23.37 & 0.06 & 0.00 & 3.36 & 9.40 & 20.13 & 12.75 & 0.04 \\
\quad \combinedcfg{CRCL} & \textbf{8.25} & \textbf{28.52} & \textbf{42.27} & \textbf{60.14} & \textbf{79.04} & \textbf{0.19} & \textbf{6.71} & \textbf{22.82} & \textbf{35.79} & \textbf{51.01} & \textbf{65.32} & \textbf{0.16} \\
IDeal Source & 5.50 & 18.56 & 29.90 & 43.99 & 53.95 & 0.14 & 6.71 & 20.81 & 36.91 & 51.68 & 64.43 & 0.15 \\
\quad \combinedcfg{IDeal Source} & \textbf{9.28} & \textbf{30.81} & \textbf{44.10} & \textbf{62.43} & \textbf{84.19} & \textbf{0.20} & \textbf{7.38} & \textbf{32.44} & \textbf{41.39} & \textbf{58.84} & \textbf{81.21} & \textbf{0.20} \\
RoMa & 5.84 & 19.93 & 30.93 & 49.83 & 56.70 & 0.15 & 4.70 & 21.48 & 34.23 & 51.01 & 60.40 & 0.15 \\
\quad \combinedcfg{RoMa} & \textbf{9.62} & \textbf{35.97} & \textbf{49.83} & \textbf{64.72} & \textbf{95.42} & \textbf{0.22} & \textbf{8.95} & \textbf{30.43} & \textbf{41.16} & \textbf{57.94} & \textbf{80.54} & \textbf{0.20} \\
\midrule
\multicolumn{13}{@{}l}{\textbf{BERT}}\\
VSE$\infty$ & 2.75 & 8.59 & 19.24 & 32.30 & 30.58 & 0.08 & 4.03 & 12.08 & 22.15 & 36.91 & 38.26 & 0.10 \\
\quad \combinedcfg{VSE$\infty$} & \textbf{8.36} & \textbf{29.78} & \textbf{42.73} & \textbf{60.48} & \textbf{80.87} & \textbf{0.19} & \textbf{6.94} & \textbf{26.40} & \textbf{37.81} & \textbf{55.48} & \textbf{71.14} & \textbf{0.18} \\
ESA & 0.34 & 1.72 & 3.44 & 6.87 & 5.50 & 0.02 & 0.67 & 3.36 & 6.71 & 13.42 & 10.74 & 0.04 \\
\quad \combinedcfg{ESA} & \textbf{8.02} & \textbf{23.71} & \textbf{38.03} & \textbf{55.10} & \textbf{69.76} & \textbf{0.17} & \textbf{6.71} & \textbf{24.16} & \textbf{34.00} & \textbf{51.01} & \textbf{64.88} & \textbf{0.16} \\
CRCL & 1.37 & 6.53 & 12.37 & 25.43 & 20.27 & 0.06 & 3.36 & 10.74 & 18.79 & 34.23 & 32.89 & 0.09 \\
\quad \combinedcfg{CRCL} & \textbf{8.71} & \textbf{26.35} & \textbf{44.10} & \textbf{57.73} & \textbf{79.15} & \textbf{0.19} & \textbf{8.05} & \textbf{25.50} & \textbf{43.18} & \textbf{54.81} & \textbf{76.73} & \textbf{0.18} \\
IDeal Source & 4.81 & 21.99 & 34.02 & 50.17 & 60.82 & 0.15 & 5.37 & 26.17 & 40.94 & 54.36 & 72.48 & 0.17 \\
\quad \combinedcfg{IDeal Source} & \textbf{9.05} & \textbf{33.91} & \textbf{46.16} & \textbf{63.46} & \textbf{89.12} & \textbf{0.21} & \textbf{9.40} & \textbf{32.44} & \textbf{44.74} & \textbf{63.98} & \textbf{86.58} & \textbf{0.21} \\
RoMa & 6.53 & 25.09 & 35.40 & 49.83 & 67.01 & 0.16 & 7.38 & 24.83 & 37.58 & 51.01 & 69.80 & 0.17 \\
\quad \combinedcfg{RoMa} & \textbf{11.68} & \textbf{32.07} & \textbf{48.22} & \textbf{69.42} & \textbf{91.98} & \textbf{0.23} & \textbf{9.17} & \textbf{31.32} & \textbf{44.74} & \textbf{59.06} & \textbf{85.23} & \textbf{0.21} \\
\midrule
\multicolumn{13}{@{}l}{\textbf{Model-specific text encoders}}\\
CLIP2CLIP & 2.06 & 8.59 & 14.09 & 24.40 & 24.74 & 0.07 & 2.01 & 5.37 & 9.40 & 19.46 & 16.78 & 0.06 \\
\quad \combinedcfg{CLIP2CLIP} & \textbf{7.90} & \textbf{23.60} & \textbf{38.60} & \textbf{55.67} & \textbf{70.10} & \textbf{0.17} & \textbf{6.26} & \textbf{21.92} & \textbf{31.99} & \textbf{48.77} & \textbf{60.18} & \textbf{0.16} \\
CrossOver & 5.84 & 23.02 & 34.02 & 48.45 & 62.89 & 0.15 & 8.72 & 25.50 & 36.24 & 52.35 & 70.47 & 0.18 \\
\quad \combinedcfg{CrossOver} & \textbf{10.31} & \textbf{33.68} & \textbf{48.68} & \textbf{67.70} & \textbf{92.67} & \textbf{0.22} & \textbf{9.62} & \textbf{29.75} & \textbf{45.19} & \textbf{60.40} & \textbf{84.56} & \textbf{0.21} \\
\bottomrule
\end{tabular}
\end{table}

\FloatBarrier
\section{The Impact of Refinement Rounds}
\label{sec:supp_camera_rounds}
The three main-comparison checkpoints, trained with $K=12$ on both datasets at $\lambda=0.50$, are reused across all round policies. Following the main paper, $r$ indexes refinement rounds, $R$ denotes their total count, and $V_s^{(r)}$ is the accepted pose sequence after round $r$. Each round recomputes geometry descriptors and pooled scene context from $V_s^{(r-1)}$. Cumulative pose offsets remain bounded relative to $V_s^{(0)}$. Proposals are processed in \initcfg{} order and accepted when position validity, a first-hit fraction of at least 0.10, and a gain greater than $10^{-12}$ in $J_{0.50}$ are satisfied.

The fixed policies stop after one or two rounds. Adaptive refinement stops after a complete round when at least 11 slots retain their poses, with a cap of ten rounds. Each policy renders and encodes only its final views. Visual pooling uses $\tau=0.03$; fusion uses direction-specific row statistics and fixed visual/base weights of 0.75/0.25. Rankings retain the full galleries and lexical tie breaking. Fused results average text towers within each family, then weight the seven families and two datasets equally.

\paragraph{Retrieval and geometric effects.}
Table~\ref{tab:supp_camera_rounds_fusion} reports the complete comparison. Two rounds achieve the highest bidirectional fused macro Rsum3, 81.95, compared with 81.53 for one round and 81.85 for adaptive refinement. Relative to two rounds, adaptive refinement raises macro text-to-scene Rsum3 by 2.92 points for the visual index and 0.15 after fusion. The scene-to-text changes are $+0.40$ and $-0.34$, respectively. The fused contrasts have paired 95\% intervals of $[-1.55,1.74]$ and $[-1.83,1.08]$ in the two directions.

Adaptive refinement reaches a median of six rounds in each dataset. Relative to two rounds, its macro geometric objective increases by $7.36\times10^{-3}$ and coverage by 0.69 percentage points. The stopping threshold ends 1,232 scene--seed runs, including 67 at round ten; the remaining 88 reach the cap with more than one accepted move. Tables~\ref{tab:supp_camera_rounds_geometry} and~\ref{tab:supp_camera_stop_rounds} report geometry and stopping behavior.

\paragraph{Paired preparation cost.}
Timing uses the same 16 scenes per dataset and three seeds, giving 96 pairs on an RTX 3090 with two pinned CPU cores. It includes geometric preparation from given \initcfg{} poses, refinement, Surfel preparation and rendering, and native batch-16 FP16 EVA encoding. Conditions rerender and re-encode their final views in balanced order; shared preparation and common early rounds contribute the same measured times to each applicable condition. Two rounds save 0.59 seconds relative to adaptive refinement.

\begin{table}[!htbp]
\centering\small\setlength{\tabcolsep}{3pt}
\caption{\textbf{Camera-refinement rounds by dataset.} Visual and fused Rsum3 average three seeds. Fused scores average seven families at $\gamma=0.75$. Prep. is in seconds.}
\label{tab:supp_camera_rounds_fusion}
\begin{tabular}{@{}llrrrrr@{}}
\toprule
& & \multicolumn{2}{c}{Text-to-scene} & \multicolumn{2}{c}{Scene-to-text} & \\
\cmidrule(lr){3-4}\cmidrule(lr){5-6}
Dataset & Policy & Visual & Fused & Visual & Fused & Prep. (s) \\
\midrule
ScanNet & One round & 89.35 & 93.49 & 72.85 & 80.99 & 2.41 \\
 & Two rounds & 90.95 & 93.54 & 73.54 & 81.48 & 2.56 \\
 & Adaptive & 91.87 & 93.32 & 74.11 & 80.68 & 3.17 \\
\midrule
3RScan & One round & 74.50 & 79.23 & 66.22 & 72.42 & 3.04 \\
 & Two rounds & 73.38 & 79.21 & 69.57 & 73.57 & 3.18 \\
 & Adaptive & 78.30 & 79.72 & 69.80 & 73.68 & 3.75 \\
\midrule
Macro & One round & 81.92 & 86.36 & 69.54 & 76.71 & 2.72 \\
 & Two rounds & 82.16 & 86.38 & 71.56 & 77.53 & 2.87 \\
 & Adaptive & 85.08 & 86.52 & 71.96 & 77.18 & 3.46 \\
\bottomrule
\end{tabular}
\end{table}

\begin{table}[!htbp]
\centering\small
\caption{\textbf{Geometric effect of continued refinement.} Coverage is the observed fraction of the scene coverage cells, expressed as a percentage. The objective is displayed as $100J_{0.50}$.}
\label{tab:supp_camera_rounds_geometry}
\begin{tabular}{@{}lrrrrrr@{}}
\toprule
& \multicolumn{3}{c}{$100J_{0.50}$} & \multicolumn{3}{c}{Coverage (\%)} \\
\cmidrule(lr){2-4}\cmidrule(lr){5-7}
Dataset & One & Two & Adaptive & One & Two & Adaptive \\
\midrule
ScanNet & 80.26 & 80.90 & 81.50 & 78.37 & 78.70 & 79.24 \\
3RScan & 80.31 & 81.22 & 82.10 & 84.96 & 85.56 & 86.41 \\
Macro & 80.28 & 81.06 & 81.80 & 81.67 & 82.13 & 82.82 \\
\bottomrule
\end{tabular}
\end{table}

\begin{table}[!htbp]
\centering\small
\caption{\textbf{Adaptive stop rounds.} Each count represents one scene and one seed. Threshold and cap counts separate the two stop reasons at round 10.}
\label{tab:supp_camera_stop_rounds}
\begin{tabular}{@{}rrrrrr@{}}
\toprule
Round & ScanNet & 3RScan & Total & Threshold & Cap \\
\midrule
1 & 0 & 0 & 0 & 0 & 0 \\
2 & 3 & 0 & 3 & 3 & 0 \\
3 & 43 & 14 & 57 & 57 & 0 \\
4 & 115 & 61 & 176 & 176 & 0 \\
5 & 169 & 76 & 245 & 245 & 0 \\
6 & 149 & 83 & 232 & 232 & 0 \\
7 & 155 & 76 & 231 & 231 & 0 \\
8 & 83 & 57 & 140 & 140 & 0 \\
9 & 54 & 27 & 81 & 81 & 0 \\
10 & 102 & 53 & 155 & 67 & 88 \\
\bottomrule
\end{tabular}
\end{table}

\FloatBarrier

\FloatBarrier
\section{The Impact of Camera Initialization}
\label{sec:dataset_budget_selection}
At ScanNet $K=12$ and 3RScan $K=6$, we compare Greedy Surface Coverage (\initcfg{}), Uniform, and five-seed Random as camera initializers. All arms share candidates, geometry, rendering, frozen EVA features and normalized LSE at $\tau=0.03$. Each renders its initial poses directly, without learned offsets or base fusion. \initcfg{} also initializes the separate learned XYZ/yaw/pitch module. These controls assess the visual evidence supplied by the initial camera set. All scores were recomputed from frozen features.

After eligibility filtering, Uniform floors equally spaced indices from zero to $M-1$ in position--azimuth--pitch order, where $M$ is the eligible count. Coverage and Random use prefixes of fixed nested sequences; Random keeps the first 12 or 6 cameras for each of five fixed seeds. All selected cameras are distinct. Random reports mean $\pm$ sample SD over seed metrics, with dataset averaging performed within seed before macro SD. These budgets were selected on the same validation galleries, so the estimates characterize the saved galleries and banks.

\begin{table}[htbp]\centering\scriptsize\setlength{\tabcolsep}{3pt}
\caption{Text-to-scene initialization ablation, ScanNet $K=12$ / 3RScan $K=6$.}\label{tab:dataset_select_referral_to_point}
\begin{tabular}{llrrrrrr}\toprule Dataset & Initializer & R@1 & R@5 & R@10 & R@20 & Rsum3 & MRR \\\midrule
ScanNet & Coverage & 9.97 & 31.27 & 45.70 & 60.82 & 86.94 & 0.21 \\
ScanNet & Uniform & 6.87 & 23.71 & 36.77 & 49.14 & 67.35 & 0.16 \\
ScanNet & Random & 7.84$\pm$1.18 & 23.09$\pm$2.49 & 34.36$\pm$1.90 & 47.56$\pm$2.31 & 65.29$\pm$5.37 & 0.16$\pm$0.01 \\
\midrule
3RScan & Coverage & 10.74 & 28.19 & 40.94 & 57.05 & 79.87 & 0.19 \\
3RScan & Uniform & 3.36 & 18.12 & 22.82 & 36.24 & 44.30 & 0.11 \\
3RScan & Random & 4.03$\pm$1.06 & 15.17$\pm$3.20 & 24.03$\pm$4.44 & 36.91$\pm$5.62 & 43.22$\pm$8.24 & 0.11$\pm$0.02 \\
\midrule
Macro & Coverage & 10.35 & 29.73 & 43.32 & 58.94 & 83.40 & 0.20 \\
Macro & Uniform & 5.11 & 20.92 & 29.79 & 42.69 & 55.82 & 0.14 \\
Macro & Random & 5.93$\pm$1.03 & 19.13$\pm$2.50 & 29.20$\pm$2.66 & 42.24$\pm$3.06 & 54.26$\pm$6.01 & 0.13$\pm$0.01 \\
\bottomrule\end{tabular}\end{table}
\begin{table}[htbp]\centering\scriptsize\setlength{\tabcolsep}{3pt}
\caption{Scene-to-text initialization ablation, ScanNet $K=12$ / 3RScan $K=6$.}\label{tab:dataset_select_point_to_referral}
\begin{tabular}{llrrrrrr}\toprule Dataset & Initializer & R@1 & R@5 & R@10 & R@20 & Rsum3 & MRR \\\midrule
ScanNet & Coverage & 7.56 & 27.84 & 40.55 & 55.33 & 75.95 & 0.18 \\
ScanNet & Uniform & 5.15 & 19.93 & 30.24 & 43.99 & 55.33 & 0.14 \\
ScanNet & Random & 5.02$\pm$0.79 & 19.45$\pm$2.07 & 30.17$\pm$2.54 & 45.36$\pm$1.46 & 54.64$\pm$4.26 & 0.13$\pm$0.01 \\
\midrule
3RScan & Coverage & 5.37 & 28.86 & 38.93 & 52.35 & 73.15 & 0.17 \\
3RScan & Uniform & 4.03 & 16.11 & 23.49 & 36.24 & 43.62 & 0.11 \\
3RScan & Random & 2.82$\pm$1.10 & 14.09$\pm$1.71 & 22.01$\pm$2.20 & 33.29$\pm$3.57 & 38.93$\pm$4.55 & 0.10$\pm$0.01 \\
\midrule
Macro & Coverage & 6.46 & 28.35 & 39.74 & 53.84 & 74.55 & 0.17 \\
Macro & Uniform & 4.59 & 18.02 & 26.87 & 40.11 & 49.48 & 0.12 \\
Macro & Random & 3.92$\pm$0.43 & 16.77$\pm$1.79 & 26.09$\pm$2.22 & 39.32$\pm$2.25 & 46.78$\pm$4.14 & 0.12$\pm$0.01 \\
\bottomrule\end{tabular}\end{table}
\paragraph{Results.}
\initcfg{} achieves macro text-to-scene Rsum3 of 83.40, compared with 55.82 for Uniform and $54.26\pm6.01$ for Random. The corresponding scene-to-text values are 74.55, 49.48, and $46.78\pm4.14$. The gains over both controls are positive in the paired analysis.

Section~\ref{sec:supp_camera_lambda_sweep} uses this fixed \initcfg{} bank as the reference for learned refinement at the same budgets.
\FloatBarrier

\FloatBarrier
\section{The Impact of Camera Supervision}
\label{sec:supp_camera_lambda_sweep}
Tables~\ref{tab:lambda_sweep_scannet} and~\ref{tab:lambda_sweep_3rscan} report standalone visual retrieval in both directions on 291 ScanNet scenes at $K=12$ and 149 3RScan scenes at $K=6$. All settings share the \initcfg{} initial poses, Surfel renderer, frozen EVA encoder, and normalized LSE with $\tau=0.03$. The coefficient $\lambda$ weights normalized observation entropy in the geometric objective; $\gamma$ denotes the separate score-fusion coefficient.

Each camera slot supplies a 157-dimensional geometry descriptor. Three $6\times8$ pooled maps encode log-normalized first-hit depth, ray-hit support, and cross-view novelty (144 values). The remaining 13 values encode normalized XYZ, sine/cosine of yaw and pitch, clearance, hit fraction, view coverage fraction, and log-scaled scene extent. The network has two width-64 local ReLU layers, masked mean scene context, a further width-64 ReLU layer, and a five-output tanh head, totaling 22,853 parameters. It predicts bounded XYZ/yaw/pitch updates with fixed roll and intrinsics.

For each $\lambda\in\{0.00,0.25,0.50,0.75,1.00\}$, three matched seeds (20261117, 20261118, 20261119) align initial weights across settings. Each dataset contributes 64 fitting and 16 diagnostic scenes. All camera models, including the uniform-$K=12$ main models, use AdamW with learning rate $10^{-3}$, weight decay $10^{-4}$, batches of eight scenes, and 30 epochs, retaining the final checkpoint. Feature standardization uses equally weighted fitting-scene moments, a standard-deviation floor of 0.05, and clipping to $[-8,8]$. Padding is masked in context pooling and regression, and the loss weights scenes equally. The five coefficients share the 11 candidate teacher actions and the acceptance-rule form; $\lambda$ changes the selected targets and objective gains. This sweep evaluates one refinement round.

\begin{table}[htbp]
\centering\small\setlength{\tabcolsep}{4pt}
\caption{ScanNet camera supervision at $K=12$. Learned rows report mean $\pm$ sample SD across three seeds. Fixed \initcfg{} is a single reference. Bold marks the highest learned Rsum3 mean in each direction.}
\label{tab:lambda_sweep_scannet}
\begin{tabular}{@{}lrrrrrr@{}}
\toprule
$\lambda$ & R@1 & R@5 & R@10 & R@20 & Rsum3 & MRR \\
\midrule
\multicolumn{7}{@{}l}{\emph{Text-to-scene}} \\
Fixed \initcfg{} & $9.97$ & $31.27$ & $45.70$ & $60.82$ & $86.94$ & $0.21$ \\
$0.00$ & $10.54\!\pm\!0.52$ & $32.07\!\pm\!0.52$ & $45.70\!\pm\!2.41$ & $61.05\!\pm\!0.40$ & $88.32\!\pm\!2.59$ & $0.22\!\pm\!0.01$ \\
$0.25$ & $11.34\!\pm\!0.69$ & $33.68\!\pm\!1.24$ & $45.59\!\pm\!0.52$ & $61.86\!\pm\!0.91$ & $90.61\!\pm\!1.43$ & $0.23\!\pm\!0.01$ \\
$0.50$ & $11.91\!\pm\!1.89$ & $34.36\!\pm\!0.60$ & $46.85\!\pm\!0.52$ & $61.86\!\pm\!0.60$ & $\mathbf{93.13}\!\pm\!1.50$ & $0.23\!\pm\!0.01$ \\
$0.75$ & $11.80\!\pm\!1.39$ & $32.42\!\pm\!0.52$ & $45.59\!\pm\!0.86$ & $61.86\!\pm\!1.57$ & $89.81\!\pm\!1.70$ & $0.23\!\pm\!0.01$ \\
$1.00$ & $12.03\!\pm\!0.60$ & $32.99\!\pm\!0.60$ & $44.79\!\pm\!0.99$ & $62.77\!\pm\!2.07$ & $89.81\!\pm\!1.43$ & $0.23\!\pm\!0.00$ \\
\midrule
\multicolumn{7}{@{}l}{\emph{Scene-to-text}} \\
Fixed \initcfg{} & $7.56$ & $27.84$ & $40.55$ & $55.33$ & $75.95$ & $0.18$ \\
$0.00$ & $7.45\!\pm\!0.40$ & $28.75\!\pm\!0.20$ & $40.21\!\pm\!0.91$ & $55.78\!\pm\!0.20$ & $76.40\!\pm\!0.99$ & $0.18\!\pm\!0.00$ \\
$0.25$ & $7.33\!\pm\!0.40$ & $29.21\!\pm\!0.91$ & $41.24\!\pm\!0.91$ & $55.67\!\pm\!0.34$ & $\mathbf{77.78}\!\pm\!2.07$ & $0.18\!\pm\!0.00$ \\
$0.50$ & $6.99\!\pm\!0.20$ & $27.84\!\pm\!0.91$ & $40.09\!\pm\!0.20$ & $56.13\!\pm\!0.52$ & $74.91\!\pm\!0.91$ & $0.18\!\pm\!0.00$ \\
$0.75$ & $6.87\!\pm\!0.60$ & $28.75\!\pm\!0.99$ & $39.18\!\pm\!0.34$ & $55.78\!\pm\!0.99$ & $74.80\!\pm\!1.55$ & $0.18\!\pm\!0.00$ \\
$1.00$ & $7.45\!\pm\!0.52$ & $28.98\!\pm\!0.99$ & $39.75\!\pm\!0.40$ & $57.04\!\pm\!0.69$ & $76.17\!\pm\!1.10$ & $0.18\!\pm\!0.00$ \\
\bottomrule
\end{tabular}
\end{table}

\begin{table}[htbp]
\centering\small\setlength{\tabcolsep}{4pt}
\caption{3RScan camera supervision at $K=6$. Learned rows report mean $\pm$ sample SD across three seeds. Fixed \initcfg{} is a single reference. Bold marks the highest learned Rsum3 mean in each direction.}
\label{tab:lambda_sweep_3rscan}
\begin{tabular}{@{}lrrrrrr@{}}
\toprule
$\lambda$ & R@1 & R@5 & R@10 & R@20 & Rsum3 & MRR \\
\midrule
\multicolumn{7}{@{}l}{\emph{Text-to-scene}} \\
Fixed \initcfg{} & $10.74$ & $28.19$ & $40.94$ & $57.05$ & $79.87$ & $0.19$ \\
$0.00$ & $9.40\!\pm\!0.67$ & $28.19\!\pm\!2.42$ & $40.94\!\pm\!1.34$ & $57.72\!\pm\!1.16$ & $78.52\!\pm\!2.68$ & $0.19\!\pm\!0.01$ \\
$0.25$ & $10.07\!\pm\!0.67$ & $28.19\!\pm\!0.67$ & $42.06\!\pm\!0.39$ & $57.49\!\pm\!1.03$ & $\mathbf{80.31}\!\pm\!1.03$ & $0.20\!\pm\!0.00$ \\
$0.50$ & $10.07\!\pm\!0.67$ & $27.74\!\pm\!2.05$ & $40.72\!\pm\!0.39$ & $57.49\!\pm\!1.94$ & $78.52\!\pm\!2.42$ & $0.19\!\pm\!0.01$ \\
$0.75$ & $9.17\!\pm\!1.40$ & $26.62\!\pm\!0.77$ & $40.94\!\pm\!0.67$ & $57.94\!\pm\!1.03$ & $76.73\!\pm\!2.36$ & $0.19\!\pm\!0.01$ \\
$1.00$ & $8.50\!\pm\!2.05$ & $26.40\!\pm\!1.03$ & $40.94\!\pm\!1.78$ & $57.05\!\pm\!0.67$ & $75.84\!\pm\!4.08$ & $0.19\!\pm\!0.02$ \\
\midrule
\multicolumn{7}{@{}l}{\emph{Scene-to-text}} \\
Fixed \initcfg{} & $5.37$ & $28.86$ & $38.93$ & $52.35$ & $73.15$ & $0.17$ \\
$0.00$ & $6.26\!\pm\!0.39$ & $27.52\!\pm\!2.42$ & $39.60\!\pm\!0.67$ & $53.24\!\pm\!0.39$ & $\mathbf{73.38}\!\pm\!2.16$ & $0.17\!\pm\!0.00$ \\
$0.25$ & $6.71\!\pm\!1.34$ & $26.40\!\pm\!0.77$ & $37.36\!\pm\!1.03$ & $51.45\!\pm\!1.03$ & $70.47\!\pm\!0.67$ & $0.17\!\pm\!0.01$ \\
$0.50$ & $5.82\!\pm\!1.03$ & $28.41\!\pm\!0.39$ & $37.58\!\pm\!1.16$ & $52.80\!\pm\!1.69$ & $71.81\!\pm\!1.16$ & $0.17\!\pm\!0.00$ \\
$0.75$ & $6.04\!\pm\!1.16$ & $27.52\!\pm\!2.01$ & $37.58\!\pm\!0.67$ & $52.80\!\pm\!1.40$ & $71.14\!\pm\!1.34$ & $0.17\!\pm\!0.01$ \\
$1.00$ & $7.16\!\pm\!0.39$ & $27.52\!\pm\!0.67$ & $38.03\!\pm\!2.05$ & $51.68\!\pm\!1.16$ & $72.71\!\pm\!2.79$ & $0.18\!\pm\!0.00$ \\
\bottomrule
\end{tabular}
\end{table}

\FloatBarrier

\paragraph{Results.}
On ScanNet, $\lambda=0.50$ gives the highest mean text-to-scene Rsum3, 93.13: a gain of 6.19 over fixed \initcfg{} (95\% interval $[1.58,11.00]$). On 3RScan, $\lambda=0.25$ gives the highest mean. The scene-to-text maxima occur at $\lambda=0.25$ on ScanNet and $\lambda=0.00$ on 3RScan. Paired intervals for contrasts between learned settings include zero.

The text-to-scene contrast between $\lambda=0.50$ and 0.25 is $+2.52$ on ScanNet (95\% interval $[-1.65,6.78]$) and $-1.79$ on 3RScan ($[-7.79,4.73]$). The scene-to-text contrast between 0.25 and 0.00 is $+1.37$ and $-2.91$, respectively. Cross-dataset differences reflect both dataset and budget ($K=12$ versus $K=6$).
\FloatBarrier

\FloatBarrier
\section{The Impact of Pose Degrees of Freedom}
\label{sec:supp_camera_pose_dof}
Tables~\ref{tab:pose_dof_scannet} and~\ref{tab:pose_dof_3rscan} report all six metrics in both directions on the 291-scene ScanNet gallery at $K=12$ and the 149-scene 3RScan gallery at $K=6$. All learned policies use $\lambda=0.50$, the same \initcfg{} initialization, Surfel rendering, frozen EVA, and normalized LSE with $\tau=0.03$.

XYZ, XYZ+yaw, and XYZ+pitch each train for 30 epochs from three matched initializations on the same 128 fitting and 32 diagnostic scenes. All four learned settings use the same five-output head. For restricted settings, the teacher probes only enabled axes, and a common mask zeros disabled displacements during training and inference. The full policy reuses the three verified $\lambda=0.50$ models from Appendix~\ref{sec:supp_camera_lambda_sweep}. Roll and intrinsics remain fixed, and every learned row uses the final checkpoint.

\begin{table}[htbp]
\centering\small\setlength{\tabcolsep}{4pt}
\caption{ScanNet camera pose ablation at $K=12$ and $\lambda=0.50$. Learned rows show mean $\pm$ sample SD over three seeds. Fixed \initcfg{} is the unrefined reference. Bold marks the highest mean Rsum3 in each direction.}
\label{tab:pose_dof_scannet}
\begin{tabular}{@{}lrrrrrr@{}}
\toprule
Pose update & R@1 & R@5 & R@10 & R@20 & Rsum3 & MRR \\
\midrule
\multicolumn{7}{@{}l}{\emph{Text-to-scene}} \\
Fixed \initcfg{} & $9.97$ & $31.27$ & $45.70$ & $60.82$ & $86.94$ & $0.21$ \\
XYZ & $10.19\!\pm\!0.86$ & $32.99\!\pm\!0.69$ & $46.51\!\pm\!0.72$ & $61.17\!\pm\!1.03$ & $89.69\!\pm\!0.69$ & $0.22\!\pm\!0.01$ \\
XYZ+yaw & $11.23\!\pm\!1.62$ & $32.42\!\pm\!0.99$ & $46.05\!\pm\!0.69$ & $61.63\!\pm\!0.20$ & $89.69\!\pm\!2.81$ & $0.22\!\pm\!0.01$ \\
XYZ+pitch & $10.77\!\pm\!0.52$ & $32.19\!\pm\!1.10$ & $47.54\!\pm\!0.79$ & $61.74\!\pm\!1.43$ & $90.49\!\pm\!1.89$ & $0.22\!\pm\!0.00$ \\
XYZ+yaw+pitch & $11.91\!\pm\!1.89$ & $34.36\!\pm\!0.60$ & $46.85\!\pm\!0.52$ & $61.86\!\pm\!0.60$ & $\mathbf{93.13}\!\pm\!1.50$ & $0.23\!\pm\!0.01$ \\
\midrule
\multicolumn{7}{@{}l}{\emph{Scene-to-text}} \\
Fixed \initcfg{} & $7.56$ & $27.84$ & $40.55$ & $55.33$ & $75.95$ & $0.18$ \\
XYZ & $7.33\!\pm\!0.52$ & $28.06\!\pm\!0.52$ & $40.21\!\pm\!0.69$ & $55.90\!\pm\!0.52$ & $75.60\!\pm\!1.19$ & $0.18\!\pm\!0.00$ \\
XYZ+yaw & $8.36\!\pm\!0.40$ & $28.87\!\pm\!1.57$ & $39.98\!\pm\!0.20$ & $56.93\!\pm\!0.86$ & $\mathbf{77.21}\!\pm\!1.39$ & $0.18\!\pm\!0.00$ \\
XYZ+pitch & $6.87\!\pm\!0.60$ & $28.29\!\pm\!0.20$ & $39.06\!\pm\!0.40$ & $56.47\!\pm\!0.86$ & $74.23\!\pm\!0.69$ & $0.17\!\pm\!0.00$ \\
XYZ+yaw+pitch & $6.99\!\pm\!0.20$ & $27.84\!\pm\!0.91$ & $40.09\!\pm\!0.20$ & $56.13\!\pm\!0.52$ & $74.91\!\pm\!0.91$ & $0.18\!\pm\!0.00$ \\
\bottomrule
\end{tabular}
\end{table}

\begin{table}[htbp]
\centering\small\setlength{\tabcolsep}{4pt}
\caption{3RScan camera pose ablation at $K=6$ and $\lambda=0.50$. Learned rows show mean $\pm$ sample SD over three seeds. Fixed \initcfg{} is the unrefined reference. Bold marks the highest mean Rsum3 in each direction.}
\label{tab:pose_dof_3rscan}
\begin{tabular}{@{}lrrrrrr@{}}
\toprule
Pose update & R@1 & R@5 & R@10 & R@20 & Rsum3 & MRR \\
\midrule
\multicolumn{7}{@{}l}{\emph{Text-to-scene}} \\
Fixed \initcfg{} & $10.74$ & $28.19$ & $40.94$ & $57.05$ & $79.87$ & $0.19$ \\
XYZ & $10.29\!\pm\!1.40$ & $27.74\!\pm\!1.03$ & $42.06\!\pm\!0.39$ & $55.93\!\pm\!1.03$ & $\mathbf{80.09}\!\pm\!2.54$ & $0.20\!\pm\!0.01$ \\
XYZ+yaw & $8.50\!\pm\!0.39$ & $28.41\!\pm\!0.77$ & $41.39\!\pm\!0.77$ & $55.70\!\pm\!1.16$ & $78.30\!\pm\!0.39$ & $0.19\!\pm\!0.01$ \\
XYZ+pitch & $9.40\!\pm\!0.67$ & $27.52\!\pm\!1.78$ & $42.06\!\pm\!1.55$ & $54.14\!\pm\!1.03$ & $78.97\!\pm\!3.70$ & $0.19\!\pm\!0.01$ \\
XYZ+yaw+pitch & $10.07\!\pm\!0.67$ & $27.74\!\pm\!2.05$ & $40.72\!\pm\!0.39$ & $57.49\!\pm\!1.94$ & $78.52\!\pm\!2.42$ & $0.19\!\pm\!0.01$ \\
\midrule
\multicolumn{7}{@{}l}{\emph{Scene-to-text}} \\
Fixed \initcfg{} & $5.37$ & $28.86$ & $38.93$ & $52.35$ & $73.15$ & $0.17$ \\
XYZ & $6.04\!\pm\!0.67$ & $27.74\!\pm\!1.40$ & $38.93\!\pm\!0.00$ & $52.13\!\pm\!1.69$ & $72.71\!\pm\!1.03$ & $0.17\!\pm\!0.00$ \\
XYZ+yaw & $6.49\!\pm\!0.77$ & $28.86\!\pm\!1.78$ & $38.03\!\pm\!1.69$ & $51.90\!\pm\!0.77$ & $\mathbf{73.38}\!\pm\!2.05$ & $0.17\!\pm\!0.00$ \\
XYZ+pitch & $5.59\!\pm\!0.77$ & $27.74\!\pm\!1.03$ & $37.81\!\pm\!0.77$ & $51.68\!\pm\!1.78$ & $71.14\!\pm\!2.42$ & $0.17\!\pm\!0.01$ \\
XYZ+yaw+pitch & $5.82\!\pm\!1.03$ & $28.41\!\pm\!0.39$ & $37.58\!\pm\!1.16$ & $52.80\!\pm\!1.69$ & $71.81\!\pm\!1.16$ & $0.17\!\pm\!0.00$ \\
\bottomrule
\end{tabular}
\end{table}

\FloatBarrier

\paragraph{Results.}
The full policy gives the highest text-to-scene Rsum3 mean on ScanNet (93.13), while XYZ gives the highest mean on 3RScan (80.09). Adding yaw to XYZ changes Rsum3 by 0.00 and $-1.79$ points, respectively; adding pitch changes it by $+0.80$ and $-1.12$. With the other angle enabled, the yaw increments are $+2.63$ and $-0.45$, and the pitch increments are $+3.44$ and $+0.22$.

For scene-to-text, XYZ+yaw gives the highest means on ScanNet (77.21) and 3RScan (73.38). Adding pitch to this policy changes Rsum3 by $-2.29$ and $-1.57$ points. The paired 95\% intervals for the angular increments and the yaw--pitch interaction include zero in both directions.
\FloatBarrier

\FloatBarrier
\section{The Impact of Geometric Acceptance}
\label{sec:supp_camera_acceptance}
We compare fixed \initcfg{}, validity-only acceptance, and the full gain-based rule at $\lambda=0.50$. The study uses the complete 291-scene ScanNet gallery at $K=12$ and 149-scene 3RScan gallery at $K=6$. Both learned conditions replay identical proposals from the same three five-dimensional checkpoints. Rendering, frozen EVA encoding, and normalized LSE with $\tau=0.03$ are shared.

Validity-only acceptance requires a geometrically valid position and a first-hit fraction of at least 0.10. The full rule also requires an increase greater than $10^{-12}$ in $J_{0.50}$, processing proposals in \initcfg{} slot order and accounting for previously accepted updates. Rejected slots retain their initial poses. Fixed \initcfg{} renders the initial camera set directly.

\begin{table}[htbp]
\centering\small\setlength{\tabcolsep}{4pt}
\caption{ScanNet inference-time acceptance ablation at $K=12$ and $\lambda=0.50$. Learned rows report mean $\pm$ sample SD across three matched seeds. Bold marks the highest Rsum3 mean in each direction.}
\label{tab:acceptance_scannet}
\begin{tabular}{@{}lrrrrrr@{}}
\toprule
Acceptance rule & R@1 & R@5 & R@10 & R@20 & Rsum3 & MRR \\
\midrule
\multicolumn{7}{@{}l}{\emph{Text-to-scene}} \\
Fixed \initcfg{} & $9.97$ & $31.27$ & $45.70$ & $60.82$ & $86.94$ & $0.21$ \\
Validity only & $11.34\!\pm\!1.79$ & $33.79\!\pm\!1.39$ & $45.70\!\pm\!1.03$ & $62.31\!\pm\!0.52$ & $90.84\!\pm\!2.62$ & $0.23\!\pm\!0.01$ \\
Full rule & $11.91\!\pm\!1.89$ & $34.36\!\pm\!0.60$ & $46.85\!\pm\!0.52$ & $61.86\!\pm\!0.60$ & $\mathbf{93.13}\!\pm\!1.50$ & $0.23\!\pm\!0.01$ \\
\midrule
\multicolumn{7}{@{}l}{\emph{Scene-to-text}} \\
Fixed \initcfg{} & $7.56$ & $27.84$ & $40.55$ & $55.33$ & $\mathbf{75.95}$ & $0.18$ \\
Validity only & $7.33\!\pm\!0.20$ & $27.72\!\pm\!0.20$ & $39.75\!\pm\!0.20$ & $56.24\!\pm\!0.72$ & $74.80\!\pm\!0.40$ & $0.18\!\pm\!0.00$ \\
Full rule & $6.99\!\pm\!0.20$ & $27.84\!\pm\!0.91$ & $40.09\!\pm\!0.20$ & $56.13\!\pm\!0.52$ & $74.91\!\pm\!0.91$ & $0.18\!\pm\!0.00$ \\
\bottomrule
\end{tabular}
\end{table}

\begin{table}[htbp]
\centering\small\setlength{\tabcolsep}{4pt}
\caption{3RScan inference-time acceptance ablation at $K=6$ and $\lambda=0.50$. Learned rows report mean $\pm$ sample SD across three matched seeds. Bold marks the highest Rsum3 mean in each direction.}
\label{tab:acceptance_3rscan}
\begin{tabular}{@{}lrrrrrr@{}}
\toprule
Acceptance rule & R@1 & R@5 & R@10 & R@20 & Rsum3 & MRR \\
\midrule
\multicolumn{7}{@{}l}{\emph{Text-to-scene}} \\
Fixed \initcfg{} & $10.74$ & $28.19$ & $40.94$ & $57.05$ & $\mathbf{79.87}$ & $0.19$ \\
Validity only & $10.07\!\pm\!1.34$ & $27.29\!\pm\!3.17$ & $42.28\!\pm\!2.01$ & $57.49\!\pm\!0.77$ & $79.64\!\pm\!2.71$ & $0.19\!\pm\!0.01$ \\
Full rule & $10.07\!\pm\!0.67$ & $27.74\!\pm\!2.05$ & $40.72\!\pm\!0.39$ & $57.49\!\pm\!1.94$ & $78.52\!\pm\!2.42$ & $0.19\!\pm\!0.01$ \\
\midrule
\multicolumn{7}{@{}l}{\emph{Scene-to-text}} \\
Fixed \initcfg{} & $5.37$ & $28.86$ & $38.93$ & $52.35$ & $\mathbf{73.15}$ & $0.17$ \\
Validity only & $6.04\!\pm\!0.67$ & $27.96\!\pm\!0.77$ & $37.36\!\pm\!1.03$ & $53.47\!\pm\!1.03$ & $71.36\!\pm\!1.03$ & $0.17\!\pm\!0.00$ \\
Full rule & $5.82\!\pm\!1.03$ & $28.41\!\pm\!0.39$ & $37.58\!\pm\!1.16$ & $52.80\!\pm\!1.69$ & $71.81\!\pm\!1.16$ & $0.17\!\pm\!0.00$ \\
\bottomrule
\end{tabular}
\end{table}

\FloatBarrier

\paragraph{Geometric and retrieval effects.}
The gain check reduces acceptance from 93.74\% to 76.20\% on ScanNet and from 92.65\% to 75.02\% on 3RScan. It increases the mean final geometric gain on both datasets (Table~\ref{tab:acceptance_effects}). Relative to validity only, text-to-scene Rsum3 changes by $+2.29$ and $-1.12$ points, while scene-to-text changes by $+0.11$ and $+0.45$. All four paired 95\% intervals include zero.

\begin{table}[htbp]
\centering\small\setlength{\tabcolsep}{6pt}
\caption{Geometric diagnostics and retrieval effects of the gain check. Geometric rows give mean $\pm$ sample SD over three seeds. Objective gains are shown as $100\Delta J_{0.50}$ to retain useful precision with two decimals. Retrieval changes compare the full rule with validity-only acceptance.}
\label{tab:acceptance_effects}
\begin{tabular}{@{}lrr@{}}
\toprule
Measure & ScanNet ($K=12$) & 3RScan ($K=6$) \\
\midrule
\multicolumn{3}{@{}l}{\emph{Acceptance rate (\%)}} \\
Validity only & $93.74\!\pm\!0.27$ & $92.65\!\pm\!0.93$ \\
Full rule & $76.20\!\pm\!0.60$ & $75.02\!\pm\!1.92$ \\
\midrule
\multicolumn{3}{@{}l}{\emph{Mean final $100\Delta J_{0.50}$}} \\
Validity only & $1.75\!\pm\!0.03$ & $1.76\!\pm\!0.09$ \\
Full rule & $1.90\!\pm\!0.03$ & $1.96\!\pm\!0.08$ \\
\midrule
\multicolumn{3}{@{}l}{\emph{Rsum3 change (points)}} \\
Text-to-scene & $+2.29$ & $-1.12$ \\
Scene-to-text & $+0.11$ & $+0.45$ \\
\bottomrule
\end{tabular}
\end{table}

\FloatBarrier

\FloatBarrier
\section{The Impact of Observation Budget}
\label{sec:budget_cost}
We compare $K\in\{6,12,18,24,30\}$ using the complete validation galleries. Coverage follows one nested greedy sequence across budgets. Budgets up to 18 contain distinct cameras, while 13 scenes use the shared padding rule at 30 slots. The validation sweep informed the 12/6 budgets used for the initialization and camera-policy ablations. The main fused comparison uses 12 views on both datasets.

\begin{table}[htbp]
\centering\small
\caption{\textbf{Observation budget, retrieval, and preparation cost.} Visual Rsum3 averages both datasets equally. Preparation is mean seconds per scene, measured at all five budgets on the same 12 scenes. Storage is the unique descriptor payload across the complete galleries.}
\label{tab:camera_budget}
\setlength{\tabcolsep}{5pt}
\begin{tabular}{@{}rrrr@{}}
\toprule
$K$ & \covercfg{} (Rsum3) & Preparation (s) & Storage (MiB) \\
\midrule
6 & 81.00 & 12.08 & 10.31 \\
12 & 81.39 & 12.42 & 20.62 \\
18 & 81.38 & 13.93 & 30.94 \\
24 & 78.00 & 14.29 & 41.20 \\
30 & 78.07 & 14.61 & 51.21 \\
\bottomrule
\end{tabular}

\end{table}

Table~\ref{tab:camera_budget} shows that adding views does not monotonically improve the visual index. Preparation rises from 12.08 to 14.61 seconds per scene across the five budgets; shared geometry and attribute processing account for much of the cost. Mean visual gallery scoring takes 0.92, 2.00, 2.80, 3.75, and 4.65 milliseconds. The corresponding single-query stage sums for the four RoMa/IDeal Source configurations are 77.87, 78.95, 79.74, 80.70, and 81.72 milliseconds, including both text encoders.

The detailed tables below report both retrieval directions and stage costs. Timing uses RTX 3090 hardware, two pinned CPU cores, and separate geometry, image-encoding, and online phases. The online timing comparison fixes $\gamma=0.50$ and batch-one text encoding across budgets; the main accuracy protocol uses $\gamma=0.75$. Preparation includes decoding, attributes, cameras, visibility, selection, rendering, encoding, and output writes, while its scope omits model initialization, upstream reconstruction, base-index construction, and audits beyond decoding.
\subsection{Complete Per-Dataset Observation Budgets}
\label{sec:complete_budget_results}
The following tables give every saved standalone Coverage result for $K\in\{6,12,18,24,30\}$, with complete 291/149-scene galleries. Recall and Rsum3 use percentage points; MRR uses a 0--1 scale. Rsum3 is calculated from unrounded R@1, R@5 and R@10. The camera-policy ablations use ScanNet $K=12$ and 3RScan $K=6$, selected on these validation results; the main fusion comparison uses $K=12$ on both datasets.

\begin{table}[htbp]\centering\small
\caption{Text-to-scene retrieval: complete Coverage budget results.}\label{tab:budget_full_referral_to_point}
\begin{tabular}{llrrrrrr}\toprule Dataset & $K$ & R@1 & R@5 & R@10 & R@20 & Rsum3 & MRR \\\midrule
ScanNet & 6 & 9.28 & 29.55 & 43.30 & 59.11 & 82.13 & 0.19 \\
ScanNet & 12 & 9.97 & 31.27 & 45.70 & 60.82 & 86.94 & 0.21 \\
ScanNet & 18 & 9.62 & 30.93 & 45.70 & 61.17 & 86.25 & 0.20 \\
ScanNet & 24 & 8.59 & 30.58 & 45.02 & 60.48 & 84.19 & 0.20 \\
ScanNet & 30 & 9.28 & 32.99 & 48.11 & 63.23 & 90.38 & 0.21 \\
\midrule
3RScan & 6 & 10.74 & 28.19 & 40.94 & 57.05 & 79.87 & 0.19 \\
3RScan & 12 & 10.07 & 26.85 & 38.93 & 56.38 & 75.84 & 0.20 \\
3RScan & 18 & 7.38 & 27.52 & 41.61 & 53.02 & 76.51 & 0.18 \\
3RScan & 24 & 8.72 & 26.17 & 36.91 & 53.69 & 71.81 & 0.18 \\
3RScan & 30 & 8.05 & 24.16 & 33.56 & 52.35 & 65.77 & 0.18 \\
\bottomrule\end{tabular}\end{table}
\begin{table}[htbp]\centering\small
\caption{Scene-to-text retrieval: complete Coverage budget results.}\label{tab:budget_full_point_to_referral}
\begin{tabular}{llrrrrrr}\toprule Dataset & $K$ & R@1 & R@5 & R@10 & R@20 & Rsum3 & MRR \\\midrule
ScanNet & 6 & 7.56 & 26.12 & 40.21 & 56.36 & 73.88 & 0.18 \\
ScanNet & 12 & 7.56 & 27.84 & 40.55 & 55.33 & 75.95 & 0.18 \\
ScanNet & 18 & 6.19 & 24.40 & 39.18 & 52.23 & 69.76 & 0.17 \\
ScanNet & 24 & 6.19 & 25.43 & 37.46 & 51.20 & 69.07 & 0.17 \\
ScanNet & 30 & 6.19 & 23.71 & 37.46 & 53.26 & 67.35 & 0.17 \\
\midrule
3RScan & 6 & 5.37 & 28.86 & 38.93 & 52.35 & 73.15 & 0.17 \\
3RScan & 12 & 5.37 & 25.50 & 38.93 & 51.01 & 69.80 & 0.16 \\
3RScan & 18 & 4.70 & 25.50 & 34.23 & 48.32 & 64.43 & 0.15 \\
3RScan & 24 & 4.70 & 22.82 & 30.87 & 46.98 & 58.39 & 0.15 \\
3RScan & 30 & 4.70 & 22.82 & 31.54 & 48.32 & 59.06 & 0.15 \\
\bottomrule\end{tabular}\end{table}
\FloatBarrier

\subsection{Measured Cost by View Budget}
\label{sec:measured_view_cost}
All five budgets were measured together on the same twelve primary scenes (six point-count quantiles per dataset), with two separate warmup and two padding-control scenes. Geometry and encoding use separate processes. Each budget renders and encodes its physical observations; shared decoding, attributes, cameras and visibility are charged once to each budget.
\begin{center}\small\setlength{\tabcolsep}{4pt}
\begin{tabular}{r|rrrrrrr}\toprule
$K$ & Raw + attr. & Camera & Vis. & Select & Render & EVA & Total \\\midrule
6 & 7.92 & 0.48 & 2.16 & 0.01 & 0.16 & 1.11 & 12.08 \\
12 & 7.92 & 0.48 & 2.16 & 0.01 & 0.29 & 1.12 & 12.42 \\
18 & 7.92 & 0.48 & 2.16 & 0.01 & 0.40 & 2.32 & 13.93 \\
24 & 7.92 & 0.48 & 2.16 & 0.02 & 0.54 & 2.26 & 14.29 \\
30 & 7.92 & 0.48 & 2.16 & 0.02 & 0.69 & 2.25 & 14.61 \\
\bottomrule\end{tabular}\end{center}
Times are mean seconds per scene with equal dataset weight. Total also includes PNG and feature writes. EVA uses native mixed FP16 with 16-image padded batches. Smaller banks occupy one batch; 18/24/30 physical images occupy two. Unique descriptor payloads are 10.31, 20.62, 30.94, 41.20, 51.21 MiB across both full galleries for 6/12/18/24/30 slots; repeats contribute separately to pooling.

Online measurements cover every query in the complete 291/149-scene galleries and all four RoMa/IDeal Source text-tower configurations. Each request encodes the query once with each text encoder and reuses the two embeddings across five paired budget evaluations. The table reports means of configuration-level quantiles in milliseconds; Visual denotes scoring against the cached visual gallery.
\begin{center}\small\setlength{\tabcolsep}{4pt}
\begin{tabular}{lr|rrrr}\toprule
Gallery & $K$ & Visual P50 & Visual P95 & Total P50 & Total P95 \\\midrule
ScanNet & 6 & 0.96 & 1.23 & 47.91 & 343.71 \\
ScanNet & 12 & 1.62 & 2.07 & 48.63 & 344.58 \\
ScanNet & 18 & 2.35 & 3.01 & 49.35 & 344.95 \\
ScanNet & 24 & 2.93 & 4.22 & 50.02 & 346.09 \\
ScanNet & 30 & 3.62 & 5.04 & 50.67 & 348.07 \\
3RScan & 6 & 0.59 & 0.76 & 50.70 & 320.65 \\
3RScan & 12 & 0.96 & 1.16 & 51.04 & 321.03 \\
3RScan & 18 & 1.36 & 1.68 & 51.67 & 331.77 \\
3RScan & 24 & 1.71 & 2.12 & 51.83 & 323.88 \\
3RScan & 30 & 2.06 & 2.61 & 52.28 & 344.46 \\
\bottomrule\end{tabular}\end{center}
The accuracy tables use the saved batch-16 exports. The batch-one timing path differs from them by at most $3.08\times10^{-4}$ and changes 5 top-1 predictions across 8,800 query/budget/configuration comparisons. Timing comparisons fix $\gamma=0.50$ across budgets. RTX 3090 device identities, sampled process occupancy and memory, two-core affinity, stage source hashes, and raw timing records are preserved. The reported timing scope omits model initialization, base-index construction, upstream reconstruction, and audits outside decoding; host and filesystem cache state follow the observed execution.

\FloatBarrier

\FloatBarrier
\section{Vision--Language Backbone Substitution}
\label{sec:supp_clip_backbone}
All backbones receive the same two-round RGB views and queries for each of three matched camera seeds on the complete ScanNet (291 scenes) and 3RScan (149 scenes) validation galleries. Both datasets use $K=12$ and $\lambda=0.50$. The visual score pools paired cosine similarities with normalized LSE at $\tau=0.03$. Fusion uses direction-specific row standardization and $\gamma=0.75$. Fused results weight the seven base families equally after averaging each family's text configurations.

We use the released
\href{https://huggingface.co/timm/eva02_enormous_patch14_plus_clip_224.laion2b_s9b_b144k/tree/0162f50c2742be3c90b3a81e4e8066fc5a90e125}{EVA02-E-14-plus}~\cite{evaclip},
\href{https://huggingface.co/laion/CLIP-ViT-bigG-14-laion2B-39B-b160k/tree/743c27bd53dfe508a0ade0f50698f99b39d03bec}{OpenCLIP ViT-bigG/14 LAION-2B}~\cite{cherti2023reproducible}, and
\href{https://huggingface.co/google/siglip2-base-patch16-224/tree/75de2d55ec2d0b4efc50b3e9ad70dba96a7b2fa2}{SigLIP 2 Base/16 at $224\times224$}~\cite{tschannen2025siglip2}
checkpoints. Each uses its paired image and text towers, native image preprocessing, and native tokenizer. The 77-token EVA and OpenCLIP windows truncate 238 ScanNet and 143 3RScan queries. The 64-token SigLIP 2 window truncates 251 and 144, respectively.

Tables~\ref{tab:supp_clip_backbone_scannet} and~\ref{tab:supp_clip_backbone_3rscan} give both retrieval directions for visual-only and fused retrieval.

\begin{table}[!htbp]
\centering\small
\caption{\textbf{Backbone substitution on ScanNet.} Means over three matched camera seeds.}
\label{tab:supp_clip_backbone_scannet}
\begin{tabular}{@{}llrrrrrr@{}}
\toprule
Backbone & Retrieval & R@1 & R@5 & R@10 & R@20 & Rsum3 & MRR \\
\midrule
\multicolumn{8}{@{}l}{\textit{Text-to-scene}} \\
EVA02-E-14-plus & Visual & 11.80 & 32.19 & 46.96 & 63.34 & 90.95 & 0.23 \\
OpenCLIP bigG/14 & Visual & 7.90 & 24.40 & 37.46 & 53.61 & 69.76 & 0.17 \\
SigLIP 2 Base/16 & Visual & 6.30 & 20.50 & 31.16 & 43.87 & 57.96 & 0.15 \\
EVA02-E-14-plus & Fused & 11.59 & 33.78 & 48.17 & 66.63 & 93.54 & 0.23 \\
OpenCLIP bigG/14 & Fused & 8.35 & 26.54 & 39.89 & 56.73 & 74.77 & 0.18 \\
SigLIP 2 Base/16 & Fused & 6.82 & 22.09 & 34.47 & 50.56 & 63.38 & 0.16 \\
\midrule
\multicolumn{8}{@{}l}{\textit{Scene-to-text}} \\
EVA02-E-14-plus & Visual & 6.87 & 26.69 & 39.98 & 57.39 & 73.54 & 0.17 \\
OpenCLIP bigG/14 & Visual & 6.07 & 20.96 & 31.62 & 47.88 & 58.65 & 0.14 \\
SigLIP 2 Base/16 & Visual & 1.95 & 12.60 & 20.85 & 31.27 & 35.40 & 0.08 \\
EVA02-E-14-plus & Fused & 8.76 & 29.10 & 43.63 & 61.09 & 81.48 & 0.20 \\
OpenCLIP bigG/14 & Fused & 6.77 & 23.47 & 36.48 & 53.06 & 66.72 & 0.16 \\
SigLIP 2 Base/16 & Fused & 4.00 & 15.64 & 24.75 & 37.65 & 44.40 & 0.11 \\
\bottomrule
\end{tabular}
\end{table}

\begin{table}[!htbp]
\centering\small
\caption{\textbf{Backbone substitution on 3RScan.} Means over three matched camera seeds.}
\label{tab:supp_clip_backbone_3rscan}
\begin{tabular}{@{}llrrrrrr@{}}
\toprule
Backbone & Retrieval & R@1 & R@5 & R@10 & R@20 & Rsum3 & MRR \\
\midrule
\multicolumn{8}{@{}l}{\textit{Text-to-scene}} \\
EVA02-E-14-plus & Visual & 8.28 & 26.85 & 38.26 & 58.17 & 73.38 & 0.19 \\
OpenCLIP bigG/14 & Visual & 3.80 & 18.79 & 33.11 & 46.98 & 55.70 & 0.13 \\
SigLIP 2 Base/16 & Visual & 5.37 & 17.45 & 30.65 & 43.40 & 53.47 & 0.13 \\
EVA02-E-14-plus & Fused & 8.61 & 28.28 & 42.31 & 59.64 & 79.21 & 0.19 \\
OpenCLIP bigG/14 & Fused & 5.93 & 22.13 & 35.01 & 50.56 & 63.07 & 0.15 \\
SigLIP 2 Base/16 & Fused & 6.39 & 21.38 & 33.62 & 46.31 & 61.39 & 0.15 \\
\midrule
\multicolumn{8}{@{}l}{\textit{Scene-to-text}} \\
EVA02-E-14-plus & Visual & 6.49 & 27.07 & 36.02 & 50.56 & 69.57 & 0.17 \\
OpenCLIP bigG/14 & Visual & 4.70 & 21.92 & 31.32 & 46.76 & 57.94 & 0.14 \\
SigLIP 2 Base/16 & Visual & 3.80 & 11.63 & 18.12 & 28.86 & 33.56 & 0.09 \\
EVA02-E-14-plus & Fused & 7.65 & 26.93 & 38.99 & 55.26 & 73.57 & 0.18 \\
OpenCLIP bigG/14 & Fused & 6.06 & 22.20 & 35.03 & 52.86 & 63.28 & 0.16 \\
SigLIP 2 Base/16 & Fused & 3.99 & 14.03 & 21.20 & 36.00 & 39.23 & 0.11 \\
\bottomrule
\end{tabular}
\end{table}

\FloatBarrier

\section{Training of Base Retrievers}
\label{sec:joint_training}
Ten models are trained jointly on ScanNet and 3RScan with one fixed seed and the original model and optimizer settings. Checkpoint selection maximizes equal-dataset validation text-to-scene Rsum3, with earliest-epoch tie breaking. Each selected checkpoint supplies full-gallery scores in both directions. Table~\ref{tab:fits} reports the completed schedules; Table~\ref{tab:presets} gives the early-stopping and optimization settings.
\begin{table}[ht]
\centering
\caption{Completed fresh fits and validation-selected epochs.}
\label{tab:fits}
\begin{tabular}{llrr}
\toprule
\basecfg{} family & Text encoder & Completed epochs & Selected epoch \\
\midrule
RoMa & BERT & 30 & 18 \\
RoMa & BiGRU & 300 & 40 \\
IDeal Source & BERT & 15 & 14 \\
IDeal Source & BiGRU & 15 & 13 \\
ESA & BERT & 13 & 3 \\
ESA & BiGRU & 40 & 30 \\
CRCL & BERT & 30 & 29 \\
CRCL & BiGRU & 300 & 30 \\
VSE$\infty$ & BERT & 30 & 29 \\
VSE$\infty$ & BiGRU & 15 & 5 \\
\bottomrule
\end{tabular}

\end{table}

The frozen training plan fixes the initial learning rate to $3\times10^{-4}$ for all ten fits. Table~\ref{tab:presets} lists the native batch, learning-rate update interval, early-stop patience and epoch ceiling. Patience zero disables early stopping. CRCL's four stages use BERT epochs 4/4/4/18 and BiGRU epochs 40/40/40/180. BiGRU vocabulary fitting uses only joint-training queries, native lower-case NLTK tokenization and occurrence threshold 4, preserving the native special tokens.
\begin{table}[ht]
\centering\small
\caption{Base-retriever training settings. LR-update intervals and epoch limits are in epochs.}
\label{tab:presets}
\begin{tabular}{lrrrrr}
\toprule
\basecfg{} family & Batch & LR update & Patience & BERT ceiling & BiGRU ceiling \\
\midrule
RoMa & 8 & 14 & 0 & 30 & 300 \\
IDeal Source & 5 & 10 & 0 & 15 & 15 \\
ESA & 8 & 14 & 10 & 30 & 300 \\
CRCL & 8 & 14 & 0 & 30 & 300 \\
VSE$\infty$ & 8 & 14 & 10 & 30 & 300 \\
\bottomrule
\end{tabular}

\end{table}

The shared frozen DGCNN uses the published 1,024-point, k20 checkpoint, with 200 local patches of 1,024 nearest neighbors per scene. Patch coordinates use native unit-sphere normalization. Descriptor sampling uses a fixed dataset/scan-ID-seeded generator. Query construction resets a fixed pseudorandom generator for every dataset/split processor, follows official split order, and preserves annotation-row order before sampling. The retrievers apply their native normalization to raw patch-center positions. The full ScanNet and 3RScan banks retain the sampled vertex IDs and scene orders. The reconstructed descriptor bank has independent hashes; its byte identity with the original RoMa descriptor bank was not established.

CLIP2CLIP uses frozen OpenAI CLIP ViT-B/32 and its original scene-scoring rule on fixed \initcfg{} Surfel images. It receives no training on target retrieval labels. CrossOver uses the released same-split checkpoint, including its auxiliary RGB/CAD training exposure. All methods share query and candidate identities while retaining these training histories.

\section{Point Geometry, Rendering, and Repeated Slots}
\label{sec:point_geometry}
Camera generation uses reconstructed XYZ in metric, Z-up coordinates, estimated horizontal support, and observed-point clearance. Occupancy is computed at resolution $h_{\rm occ}$; Eq.~\ref{eq:visibility_definition} defines first-hit visibility and coverage cells. The first-hit ray parameter supplies camera-Z depth. Surfel rendering uses reconstructed RGB and estimated normals, with 30 neighbors for normal estimation and orientation. For each point, local spacing $d_i$ is the median distance to six distinct neighboring points, in meters. With scene spacing $d_{\mathrm{scene}}=\operatorname{median}_i d_i$, the metric Surfel radius is $r_i=\operatorname{clip}(1.50d_i,0.50d_{\mathrm{scene}},3d_{\mathrm{scene}})$. Coincident input points retain the spacing of their unique coordinate. The 3RScan color decoder uses the release's triangle/material/texture association to obtain vertex RGB.

Two input-compatibility cases are handled before retrieval scoring. An indexed-UV decoder assigns colors to 77 vertices outside all triangle faces in 37 scenes after checking equal vertex/UV counts and identical vertex/UV face indices under one textured material. For one scene whose floor support spans neighboring height bins, a symmetric support band with half-width $h_b$ supplies 96 cells and two camera positions, satisfying the 80-cell requirement. The geometry bank covers all 157 official 3RScan validation scans; retrieval uses the 149 scans with paired referrals.

The main comparison uses 12 slots per scene, and the selector-only and camera-policy comparisons use 12 ScanNet or 6 3RScan slots. The observation-budget ablation evaluates 6, 12, 18, 24, and 30 slots; when a requested budget exceeds the eligible-camera count, a fixed scene-specific padding sequence supplies repeated identities that count individually in normalized LSE. Coverage and Random use sequence prefixes, while Uniform recomputes its evenly spaced eligible indices at each K. All 6/12-view initial sets contain unique camera identities. Random uses five fixed seeds.

EVA uses native 224-pixel preprocessing, 1,024-dimensional unit image/text features, a 77-token context, and batch-16 mixed-FP16 encoding. Text scoring places the unused visual tower on CPU; the encoder parameters and all 440 text vectors match the reference hashes under this placement. Each branch receives the complete query string and applies its native tokenizer and context limit.

\subsection{Camera and Visibility Parameters}
Equations~\ref{eq:floor_estimator}--\ref{eq:candidate_pose_set} map each reconstructed point cloud to its ordered candidate cameras. They compute the floor from local horizontal support, filter grid centers by support and clearance, apply deterministic farthest-point selection, and pair the retained centers with the fixed orientation set. Rendering uses $640\times480$ images with focal lengths 420 and principal point $(320,240)$.

Equation~\ref{eq:visibility_definition} computes first-hit observations on the $160\times120$ ray grid, applies the eligibility threshold, and maps hit-cell centers to coverage cells. Camera order is position, increasing azimuth, and the pitch order in Eq.~\ref{eq:candidate_pose_set}. Greedy ties use that order.

Algorithm~\ref{alg:c4_initialization} gives the full initialization procedure.
\FloatBarrier
\section{Implementation and Measurement Conventions}
\subsection{Coordinate and Rasterization Conventions}
\label{sec:camera_implementation_conventions}

\paragraph{Input and camera coordinates.}
ScanNet supplies reconstructed PLY vertex XYZ/RGB; 3RScan supplies reconstructed
OBJ vertices with material and texture colors. The readers preserve vertex
order and coincident vertices, without alignment, axis permutation, centering,
or unit conversion. The protocol treats the supplied coordinates as metric
and Z-up and does not estimate gravity. Coordinates are read as float32 and
promoted to float64 for geometry.
For camera center $e$, yaw $\psi$, and pitch $\theta$, define
\[
\mathbf f_v=(\cos\psi\cos\theta,\sin\psi\cos\theta,\sin\theta)^\top,\quad
\mathbf r_v=\frac{\mathbf f_v\times(0,0,1)^\top}{\|\mathbf f_v\times(0,0,1)^\top\|},\quad
\mathbf d_v=\mathbf f_v\times\mathbf r_v.
\]
The camera-to-world rotation is $R_v=[\mathbf r_v\ \mathbf d_v\ \mathbf f_v]$, so
$x_{\rm world}=R_v x_{\rm camera}+e$, with $e=t_v$ in the main paper's pose notation.
Camera axes are right, down, forward.
Zero azimuth faces world $+X$, positive azimuth turns toward $+Y$, and
positive pitch points upward. These formulas fix roll.

\paragraph{Floor and admissible camera centers.}
Let $\mathcal C_s$ contain the mean reconstructed point in each
$h_f$-voxel. A 20-neighbor PCA at $c_i\in\mathcal C_s$ gives eigenvalues
$\lambda_{i,1}\leq\lambda_{i,2}\leq\lambda_{i,3}$ and the normal $n_i$
associated with $\lambda_{i,1}$. The fixed metric constants are
\begin{equation}
\begin{aligned}
(h_f,h_b,g,h_{\rm cam},r_{\rm sup},c_{\min},d_{\min})
={}&(0.05,0.05,0.50,1.30,0.30,0.20,0.75)\,\mathrm m,\\
(h_{\rm occ},h_{\rm cov},z_{\rm near})
={}&(0.05,0.10,0.05)\,\mathrm m,
\qquad \rho_{\min}=0.10.
\end{aligned}
\label{eq:camera_metric_constants}
\end{equation}
The horizontal candidates and the scene-specific floor estimate are
\begin{equation}
\begin{aligned}
\mathcal H_s={}&\left\{c_i:\ |n_{i,z}|\geq0.95,
\frac{\lambda_{i,1}}{\max(\sum_{j=1}^{3}\lambda_{i,j},10^{-15})}\leq0.05,
c_{i,z}\leq\operatorname{median}_{c\in\mathcal C_s}c_z\right\},\\
N_b={}&\sum_{c\in\mathcal H_s}
\mathbf 1\!\left[\left\lfloor c_z/h_b\right\rfloor=b\right],\\
b_s^\star={}&\min\left\{b:N_b\geq
\max\!\left(80,0.20\max_{b'}N_{b'}\right)\right\},\\
\widehat z_s={}&\operatorname{median}\left\{c_z:c\in\mathcal H_s,
\left\lfloor c_z/h_b\right\rfloor=b_s^\star\right\}.
\end{aligned}
\label{eq:floor_estimator}
\end{equation}
With $\mathcal S_s=\{c\in\mathcal H_s:|c_z-\widehat z_s|\leq h_b\}$,
the grid and admissible centers are
\begin{equation}
\begin{aligned}
 x_i&=x_{\min}+(i+\tfrac12)g,\qquad
 y_j=y_{\min}+(j+\tfrac12)g,\\
 \mathcal I_s&=\{i\in\mathbb Z_{\geq0}:x_i<x_{\max}\},\qquad
 \mathcal J_s=\{j\in\mathbb Z_{\geq0}:y_j<y_{\max}\},\\
\mathcal G_s&=\{(x_i,y_j,\widehat z_s+h_{\rm cam}):
i\in\mathcal I_s,\ j\in\mathcal J_s\},\\
\mathcal E_s={}&\left\{e\in\mathcal G_s:
\sum_{c\in\mathcal S_s}\mathbf 1[\|c_{xy}-e_{xy}\|_2\leq r_{\rm sup}]\geq6,
\min_{(x,a)\in P_s}\|x-e\|_2\geq c_{\min}\right\}.
\end{aligned}
\label{eq:admissible_centers}
\end{equation}
Let $\bar e_{s,xy}=|\mathcal E_s|^{-1}\sum_{e\in\mathcal E_s}e_{xy}$.
The ordered farthest-point sequence is
\begin{equation}
\begin{aligned}
e_1&=\arg\min_{e\in\mathcal E_s}\|e_{xy}-\bar e_{s,xy}\|_2,\\
e_\ell&=\arg\max_{e\in\mathcal E_s\setminus\{e_1,\ldots,e_{\ell-1}\}}
\min_{k<\ell}\|e_{xy}-e_{k,xy}\|_2,\qquad \ell>1.
\end{aligned}
\label{eq:fps_centers}
\end{equation}
We retain $e_1$ and then retain $e_\ell$ only while its maximized minimum
distance is at least $d_{\min}$, stopping after at most six centers. Denote
this ordered retained set by $\widetilde{\mathcal E}_s$. Grid order breaks
exact ties.

\paragraph{Candidate orientations.}
The orientation and candidate-pose sets are
\begin{equation}
\mathcal A=\left\{\frac{2\pi j}{8}:j=0,\ldots,7\right\},\qquad
\mathcal B=\left\{-\frac{\pi}{6},-\arctan(0.15),\frac{\pi}{12}\right\},
\qquad
\mathcal P_s^{\rm cand}=\widetilde{\mathcal E}_s\times\mathcal A\times\mathcal B.
\label{eq:candidate_pose_set}
\end{equation}
In degrees, the orientation set comprises eight azimuths at $45^\circ$
intervals and pitches of $-30^\circ$, approximately $-8.53^\circ$, and
$15^\circ$. Candidate order is retained position, increasing azimuth, then the
displayed pitch order.

\paragraph{Metric and angular pose updates.}
For $\widehat{\mathbf a}\in[-1,1]^5$, define
\begin{equation}
\begin{aligned}
\mathbf b_t&=(0.25,0.25,0.125)\,\mathrm m,
&\mathbf b_r&=(\pi/18,\pi/18),\\
\Delta e&=\mathbf b_t\odot\widehat{\mathbf a}_{1:3},
&(\Delta\psi,\Delta\theta)&=\mathbf b_r\odot\widehat{\mathbf a}_{4:5},\\
\widetilde e&=e+\Delta e,
&\widetilde\theta&=\operatorname{clip}(\theta+\Delta\theta,-4\pi/9,4\pi/9),\\
\widetilde{\boldsymbol\delta}&=
(\widetilde e-e^{(0)},\operatorname{wrap}_{\pi}(\psi+\Delta\psi-\psi^{(0)}),
\widetilde\theta-\theta^{(0)}),\\
\boldsymbol\delta^\star&=\operatorname{clip}
(\widetilde{\boldsymbol\delta},-(\mathbf b_t,\mathbf b_r),(\mathbf b_t,\mathbf b_r)).
\end{aligned}
\label{eq:pose_unnormalization}
\end{equation}
Here $(e,\psi,\theta)$ is the current pose, superscript $(0)$ denotes its
initial \initcfg{} pose, $\operatorname{wrap}_{\pi}$ returns the signed
circular difference, and $\operatorname{clip}$ acts componentwise. The final
proposal reconstructs XYZ/yaw/pitch from the initial pose and
$\boldsymbol\delta^\star$. The yaw and pitch bounds are each $10^\circ$, and
the valid pitch interval is $[-80^\circ,80^\circ]$. Roll and intrinsics remain
fixed; restricted pose ablations set their disabled components to zero.

\paragraph{Grids, rays, and depth.}
Point cells use componentwise $\lfloor x/h\rfloor$, including negative
coordinates, with the world origin as grid origin. Occupancy and coverage use
the distinct fixed resolutions in Eq.~\ref{eq:camera_metric_constants}.
Visibility uses integer pixel centers on the $160\times120$ grid, without a
half-pixel offset. The visibility intrinsics are
$A_v^{\rm vis}=\operatorname{diag}(1/4,1/4,1)A_v$, with focal lengths 105
and principal point $(80,60)$. For pixel $(u,w)$, let
$\mathbf a=(A_v^{\rm vis})^{-1}(u,w,1)^\top$; the ray is
$e+tR_v\mathbf a$ without normalization. Since $a_z=1$, $t$ measures camera-Z depth.
Digital differential analyzer (DDA) traversal starts at the later of entry into
the occupied-grid axis-aligned bounding box and $z_{\rm near}$, and terminates
at the box exit. For the ray set $\mathcal R_v$, let
$q_v(\xi)$ be the center of the first occupied cell hit by ray $\xi$, or $\varnothing$ when the
ray has no hit. Eligibility and the visible coverage set are
\begin{equation}
\begin{aligned}
\rho(v)&=\frac{1}{|\mathcal R_v|}\sum_{\xi\in\mathcal R_v}
\mathbf 1[q_v(\xi)\neq\varnothing],\qquad
\mathcal P_s=\{v\in\mathcal P_s^{\rm cand}:\rho(v)\geq\rho_{\min}\},\\
C_{s,v}&=\left\{\left\lfloor\frac{q_v(\xi)}{h_{\rm cov}}\right\rfloor:
\xi\in\mathcal R_v,\ q_v(\xi)\neq\varnothing\right\}.
\end{aligned}
\label{eq:visibility_definition}
\end{equation}
The eligible pose pool $\mathcal P_s$ is the set used in the main paper's
coverage objective and Algorithm~\ref{alg:c4_initialization}; $\mathcal E_s$
contains admissible position centers. The near plane sets the lower traversal bound.
Boundary entry uses \path{nextafter} in the ray direction; simultaneous
crossings advance all axes within $10^{-12}\max(1,|t_{\rm next}|)$.
The first occupied cell is a hit.

\paragraph{Selection and error handling.}
Greedy ties use the first eligible candidate index. Selection continues to
the unique-view budget even at zero marginal gain. For $0<M<K$ eligible
cameras, all $M$ distinct cameras are retained and a shared sequence samples
$K-M$ eligible indices uniformly with replacement. Its NumPy generator is
derived deterministically from a fixed namespace, the scene ID, and the
padding purpose. The same scene-specific sequence is shared by all policies.
An empty pool, invalid XYZ, or absent supported/admitted camera position raises
an error. The floor compatibility branch catches only the
insufficient-supported-peak error: it counts the existing $\pm h_b$ band
around occupied-bin lower edges, selects the lowest qualifying band, and
rechecks support about its median. The count threshold remains
$\max(80,0.20n_{\max})$, with $n_{\max}$ now the largest band count.

\paragraph{Colors and Surfel rendering.}
For a 3RScan texture of size $W\times H$, UV maps to
$\operatorname{round}((u(W-1))\bmod W)\bmod W$ and
$\operatorname{round}(((1-v)(H-1))\bmod H)\bmod H$, using NumPy rounding.
Pillow decodes RGB; each original vertex receives the rounded mean of its
incident-corner colors, without area weighting. Untextured materials use
declared diffuse RGB. Unreferenced vertices use indexed UV only after checking
one used textured material, equal vertex/UV counts, and equal face vertex/UV
indices. Otherwise decoding fails. Source hashes are checked before and after
decoding; mesh faces serve color extraction only.

Open3D uses 30-neighbor normal estimation with
\path{fast_normal_computation=False}, then
\path{orient_normals_consistent_tangent_plane(k=30)}, with a fixed seed and
unit normalization. Integer-pixel rays intersect circular tangent
discs with the previously specified radii. Conservative near-clipped projected
bounds preserve discs whose centers are outside the image or near plane.
Rendering retains both surface orientations, without back-face or cosine culling. Intersections require camera-Z at least
$z_{\rm near}$ and squared radial distance at most $r_i^2+10^{-12}$; ray/normal dot
products below $10^{-14}$ in absolute value are skipped. Minimum depth wins;
exact ties use the lowest original vertex index. Its RGB is copied without
lighting, alpha blending, or radiometric linearization. Uncovered pixels are
$(230,230,230)$ with zero depth. Geometry/depth comparisons use float64;
outputs use uint8 RGB and float32 depth. CUDA separates minimum-depth and
minimum-index reductions, without fast math.

\subsection{Timing Sampling and Scope}
\label{sec:timing_implementation_conventions}

\paragraph{Offline measurement.}
Scenes are sorted by raw vertex count, then qualified ID. The six primary
samples per dataset occupy zero-based indices
$\lfloor(2i+1)N/12\rfloor$, $i=0,\ldots,5$. One other scene per dataset
warms all five budgets before the primary scenes; another checks padding.
The 60 primary rows exclude warmups and padding controls and contain one
measurement per scene/budget. Decoding, attributes,
cameras, and visibility run once per scene and are charged once to each
budget's stage sum. Each budget separately measures selection, rendering,
PNG writing, encoding, and feature writing. Geometry and encoding are separate
phases. Both phases use the fixed $6\rightarrow12\rightarrow18\rightarrow24\rightarrow30$ order.
EVA uses mixed FP16
batches of 16; final batches repeat the last image and discard padded outputs.

\paragraph{Online measurement.}
Each base configuration/dataset warms the first eight saved queries through
all budgets, then measures all 291/149 queries once in a fixed pseudorandom
permutation. Warmup queries also appear in
that measured pass. The two text embeddings and base scores are shared
across paired budgets; budget order cycles through rotations of $(6,12,18,24,30)$.
Each measured request produces five paired rows, totaling 8,800 query/budget
comparisons. Galleries reside on CPU. Total latency sums text,
exhaustive scoring, normalization/fusion, and sorting stages, excluding model
loading, index construction, and numerical audits. Timed fusion fixes
$\gamma=0.50$ and $\tau=0.03$ as measurement conditions. The main accuracy table uses $\gamma=0.75$.

\paragraph{Synchronization and measurement scope.}
Timers use monotonic \path{perf_counter_ns}; visibility, rendering,
image encoding, and online text blocks synchronize CUDA before stopping.
Outputs return to the host inside those blocks. Decoder integrity reads are
included in decoding time; separate image-integrity reads precede encoding.
Save-operation timing omits per-file \path{fsync}, and filesystem cache state
follows the observed execution. Reported latencies therefore cover buffered
writes under that cache state. The run uses two pinned CPU cores, one OpenBLAS
thread from process startup, and two Torch threads. Measurements use allocated
RTX 3090 devices under observed host load, fixed offline order, and one trial
per item; scene/query quantiles describe this execution rather than repeated-run
variance.

\section{Provenance and Reproduction}
The score records contain dataset-qualified query and scene IDs, the physical-source grouping, raw text hashes, and full-gallery score matrices. Image records identify the candidate cameras, retained slot sequence, repeated poses, encoder version, and normalized-LSE temperature. Checkpoint receipts bind the native model settings and exported scores. Each completed study has input and code hashes and a separate full-sort replay.

Independent replay verifies camera decisions, full-gallery rankings, summary metrics, and paired intervals before table export. Fixed-\initcfg{} controls reproduce their reference features, scores, and ranks. Table~\ref{tab:version_index} records model identities and software versions from the execution environments. The reproducibility records bind input identities, training settings, selected checkpoints, camera banks, and raw score matrices to content hashes.
OpenAI Codex helped edit or refine author-produced material for manuscript writing, related work and citations, code implementation, and experimental design, execution, and result analysis.
\par\FloatBarrier
\begin{table}[htbp]
\centering\small
\caption{Model identities and software versions. Hashes are 16-character SHA256 prefixes; the reproducibility records retain complete hashes.}\label{tab:version_index}
\begin{tabular}{>{\raggedright\arraybackslash}p{.26\linewidth}>{\raggedright\arraybackslash}p{.67\linewidth}}
\toprule
Component & Version or source identity \\
\midrule
EVA-CLIP & EVA02-E-14-plus, LAION2B, model SHA256 prefix 1e215717b06f1cb6. \\
CLIP2CLIP & CLIP ViT-B/32, checkpoint SHA256 prefix 40d365715913c9da. \\
CrossOver & Official ScanNet+3RScan weights, SHA256 prefix 125fed5ca995af0c. \\
Frozen point backbone & DGCNN checkpoint SHA256 prefix f366f60ca9dacf42. \\
Base feature/fitting environment & Python 3.9.0, PyTorch 2.7.0+cu118, NumPy 1.26.4, Transformers 4.57.6, Open3D 0.19.0. \\
Native CrossOver point environment & Python 3.10.20, PyTorch 2.7.0+cu118, NumPy 1.26.4, Open3D 0.19.0. \\
\bottomrule
\end{tabular}
\end{table}

\FloatBarrier

\end{document}